\documentclass[10pt,twocolumn,letterpaper]{article}

\usepackage[pagenumbers]{cvpr} 

\usepackage{multirow}
\definecolor{cvprblue}{rgb}{0.21,0.49,0.74}
\usepackage[pagebackref,breaklinks,colorlinks,allcolors=cvprblue]{hyperref}

\usepackage{xcolor}
\usepackage{booktabs}

\newcommand{\best}[1]{\textbf{\textcolor{red!80!black}{#1}}}
\newcommand{\secondbest}[1]{\textbf{\textcolor{blue!70!black}{#1}}}

\def\confName{CVPR}
\def\confYear{2027}

\title{SpectralRoute: Adaptive Spectral Routing for Training-Free \\Open-Vocabulary Hyperspectral Segmentation}

\author{
Qingmei Li$^{1,}$\thanks{Equal contribution,\quad $^{\dagger}$Corresponding author.}, Xiaoya Fan$^{2,*}$, Juepeng Zheng$^{3,4,\dagger}$, Jianxi Huang$^{5,6,7}$, Haohuan Fu$^{1,4}$\\
$^{1}$Tsinghua Shenzhen International Graduate School \hspace{0.2em}
$^{2}$Southwest University \hspace{0.2em}
$^{3}$Sun Yat-sen University\\
$^{4}$National Supercomputing Center in Shenzhen \quad
$^{5}$Southwest Jiaotong University \\
$^{6}$China Agricultural University \quad
$^{7}$Guangxi University of Science and Technology\\
{\tt\small qingmeili@sz.tsinghua.edu.cn, fanxiaoya@email.swu.edu.cn, zhengjp8@mail.sysu.edu.cn}
}

\usepackage{float}
\usepackage{xcolor}
\usepackage{graphicx}
\usepackage{subcaption}
\begin{document}
\maketitle

\begin{abstract}
Open-vocabulary semantic segmentation enables the   recognition of arbitrary categories through natural language descriptions and has recently been extended from natural images to remote sensing. However, existing approaches primarily focus on three-channel optical imagery or limited-band multispectral data, while pretrained vision-language models (VLMs) struggle to directly exploit the rich spectral information of hyperspectral imagery (HSI) without additional adaptation. More importantly, different land-cover categories rely on distinct spectral cues, which may be weakened by uniform fusion. To address these challenges, we propose \textbf{SpectralRoute}, a training-free framework for open-vocabulary hyperspectral segmentation. Specifically, we construct complementary spectral views from HSI and enrich category descriptions with external spectral knowledge, enabling frozen vision-language models to extract diverse spectral-semantic responses. We then introduce a class-adaptive spectral routing mechanism that selectively amplifies discriminative spectral responses for each category, overcoming the limitations of uniform fusion. Furthermore, physical spectral priors are incorporated to refine predictions for visually similar land-cover categories. Experiments on three hyperspectral benchmarks demonstrate that SpectralRoute consistently outperforms state-of-the-art methods without additional training.

\end{abstract}

\begin{figure}[t]
    \centering
    \includegraphics[width=\columnwidth]{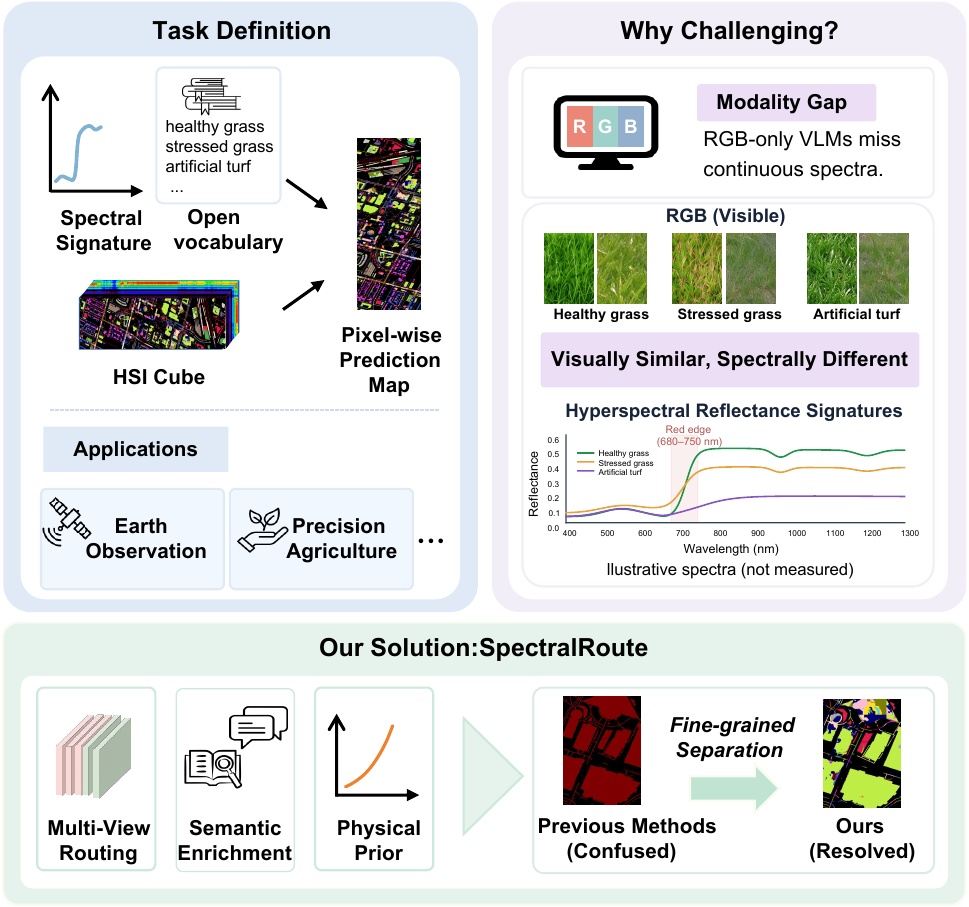}
    \vspace{-2em}
    \caption{Motivation of SpectralRoute. The task requires pixel-wise prediction from HSIs and open vocabularies. However, RGB-pretrained VLMs miss continuous spectra, causing visual ambiguity among certain categories. In contrast, hyperspectral signatures provide distinct physical evidence. SpectralRoute exploits this through multi-view routing, semantic enrichment, and physical priors to achieve fine-grained separation.}
    \label{fig:teaser}
    \vspace{-2.5em}
\end{figure}

\section{Introduction}
Hyperspectral imagery (HSI) captures detailed spectral information across numerous contiguous wavelength bands, enabling fine-grained characterization of material composition and vegetation conditions, making it particularly valuable for precision agriculture, urban monitoring, and environmental analysis~\cite{akewar2025integration, vaddi2025hyperspectral, spectralzero2026}. Despite substantial advances in hyperspectral classification and semantic segmentation, existing methods predominantly operate under a closed-set assumption~\cite{scheirer2013openset}, where models are trained to recognize a predefined set of categories using annotated samples. Such dependence limits their flexibility in handling previously unseen categories~\cite{ji2024openset, fang2024openworld} and adapting to diverse Earth observation scenarios~\cite{li2025segearth}. In practical applications, land cover categories may vary across regions, sensors, and requirements, making repeated data annotation and model retraining impractical~\cite{ye2025aaai}. Therefore, enabling flexible, text-driven recognition of land cover categories without additional training represents an important direction for hyperspectral image interpretation~\cite{wang2025tasanet, huang2025hzscm}.

Recent advances in vision-language models (VLMs), particularly CLIP~\cite{radford2021clip}, have enabled open-vocabulary semantic segmentation (OVSS)~\cite{zhou2022maskclip, xu2022simple}, allowing models to recognize categories specified by natural language descriptions. While this paradigm has been extended to remote sensing with promising results on RGB data~\cite{ye2026ldseg, Huang_2026_ICASSP, ji2026movseg}, existing methods primarily focus on RGB or limited multispectral inputs, leaving the combination of rich hyperspectral information and flexible language-defined categories insufficiently explored. Directly applying RGB-pretrained VLMs to HSIs is challenging because their visual encoders expect three-channel inputs~\cite{li2025hyperfree}. Reducing HSIs to a single RGB representation discards potentially useful spectral information, whereas incorporating additional spectral channels through learned adaptation requires further optimization~\cite{Chatterjee_2026_CVPR}. To the best of our knowledge, a dedicated framework for open-vocabulary hyperspectral semantic segmentation has not yet been established.

A natural approach to bridging this gap is to transform HSIs into multiple visual representations~\cite{zhang2026hmbench} that capture complementary cues from visible, red-edge, and near-infrared observations, alongside external spectral knowledge to enrich category names. However, the availability of multiple spectral views does not necessarily guarantee their effective utilization. A fundamental characteristic of hyperspectral imagery is that the discriminative value of spectral information varies across semantic categories~\cite{ghamisi2019multisource}. For instance, roads and buildings may be distinguished through spatial appearance, whereas vegetation conditions often exhibit informative differences in the red-edge and near-infrared regions~\cite{clevers2012using}. Similarly, grass and artificial turf can appear visually similar in RGB images while exhibiting distinct spectral responses~\cite{spectralzero2026}. These examples indicate that categories require different spectral evidence, making it difficult for a single visual representation or a shared fusion strategy to capture all relevant information.

Nevertheless, conventional multiview fusion strategies commonly aggregate semantic responses through uniform averaging or globally shared weights, overlooking the varying importance of spectral views for individual categories~\cite{fu2024clmgnet}. As a result, highly discriminative responses from informative views may be weakened by less relevant information, leading to ambiguous predictions for fine-grained categories. Furthermore, semantic similarities estimated by RGB-pretrained VLMs do not necessarily reflect the underlying physical differences between visually similar materials~\cite{Bianchi_2024_CBMI, spectralzero2026}, which can further amplify category confusion~\cite{Yang_2026_CVPR}. These limitations reveal two closely related challenges: how to selectively exploit the most informative spectral views, and how to incorporate physical spectral information for fine-grained discrimination. The key is to emphasize \textbf{category-specific spectral evidence} while maintaining \textbf{physically meaningful predictions}.

To address these challenges, we propose \textbf{SpectralRoute}, a training-free framework (Fig.~\ref{fig:teaser}). We first transform HSIs into multiple complementary visual views. In parallel, an external spectral knowledge mapping module expands simple category names into informative physical descriptions. The resulting views and descriptions are encoded by frozen CLIP encoders to establish semantic responses. Building upon these, we introduce a Joint Spectral Refinement module that combines routing and physical refinement. Specifically, Class-Adaptive Spectral Routing dynamically amplifies informative spectral evidence for each category, alleviating the loss of discriminative information caused by uniform fusion. Furthermore, Physics-Guided Refinement incorporates physically meaningful spectral cues to refine predictions for visually similar categories. The entire framework operates without target-domain annotations or additional model training.

Our main contributions are summarized as follows:
\begin{itemize}
    \item We introduce a dedicated framework for open-vocabulary hyperspectral semantic segmentation and propose SpectralRoute, which combines complementary spectral representations with knowledge-enhanced textual descriptions. It enables frozen VLMs to perform flexible pixel-level recognition without target-domain adaptation.
    \item We propose a spectral routing mechanism to address the limitations of uniform multiview fusion. By dynamically amplifying informative spectral responses for individual categories, it preserves discriminative evidence without introducing additional trainable parameters.
    \item We develop a physics-guided prediction refinement strategy that integrates physical spectral cues to reduce confusion among similar categories. We conduct systematic experiments on spectral views, individual components, parameter sensitivity, and open-vocabulary generalization.
\end{itemize}

\section{Related Work}

\subsection{Vision-Language Models}
\label{sec:vlm}

Vision-language models (VLMs) have significantly advanced multimodal understanding by learning transferable associations between visual and textual representations. Representative models, such as CLIP~\cite{radford2021clip}, ALBEF~\cite{li2021albef}, and SigLIP 2~\cite{tshannen2025siglip2}, establish effective cross-modal alignment and enable flexible zero-shot recognition. Meanwhile, generative models, including BLIP~\cite{li2022blip}, LLaVA~\cite{liu2023visual}, and MiniGPT-4~\cite{zhu2024minigpt}, further extend vision-language learning toward image captioning, instruction following, and multimodal reasoning. In remote sensing, RemoteCLIP~\cite{liu2024remoteclip}, GeoRSCLIP~\cite{zhang2024georsclip}, and SkyCLIP~\cite{wang2024skyscript} improve visual-semantic alignment through domain-specific pretraining, while SkySense-O~\cite{zhu2025skysenseo} further explores fine-grained open-world understanding of Earth observation imagery. Despite these advances, most VLMs primarily emphasize image-level semantic correspondence, which does not inherently provide accurate spatial localization or pixel-level discrimination~\cite{zhou2022maskclip}.

\subsection{Open-Vocabulary Semantic Segmentation}
\label{sec:ovss}

Open-vocabulary semantic segmentation (OVSS) extends vision-language alignment to dense prediction, enabling pixel-level recognition of categories specified by natural language. Early approaches, including MaskCLIP~\cite{zhou2022maskclip}, GroupViT~\cite{xu2022groupvit}, and TCL~\cite{cha2023tcl}, established the foundations for language-guided segmentation. Subsequent methods, such as SCLIP~\cite{wang2024sclip}, ClearCLIP~\cite{lan2024clearclip}, and ProxyCLIP~\cite{lan2024proxyclip}, improve the spatial discriminability of pretrained visual representations, while ReME~\cite{xuan2025reme} further explores training-free semantic segmentation. In remote sensing, SegEarth-OV~\cite{li2025segearth}, TMPA~\cite{Yang_2026_CVPR}, ScaleMamba~\cite{Huang_2026_ICASSP}, and RSKT-Seg~\cite{li2026rskt} address challenges associated with spatial localization, semantic ambiguity, scale variation, and domain-specific characteristics. Recent approaches, including ReAttnCLIP~\cite{Niu_2026_CVPR} and ConInfer~\cite{Chen_2026_CVPR}, further investigate training-free open-vocabulary segmentation, while KG-OVRSeg~\cite{huang2025kgovrseg} explores knowledge-enhanced semantic representations. However, existing methods predominantly focus on natural RGB imagery, with limited consideration of the rich spectral information available in HSI.

\subsection{Vision-Language Learning for Hyperspectral Imagery}
\label{sec:hsi_vl}

Hyperspectral segmentation has evolved from spectral-spatial architectures, such as 3D-CNNs~\cite{chen2016deep} and SpectralFormer~\cite{hong2022spectralformer}, to large-scale pretrained models, including SpectralGPT~\cite{hong2024spectralgpt} and HyperSIGMA~\cite{Wang_2025_HyperSIGMA}. Although these approaches have substantially advanced spectral representation learning, downstream recognition remains dependent on predefined categories or task-specific adaptation. Recent studies have begun exploring language-driven hyperspectral recognition. For instance, SPECIAL~\cite{pang2026special} combines CLIP-based pseudo-label generation with spectral learning, while HZSCM~\cite{huang2025hzscm} investigates zero-shot hyperspectral classification. Related efforts, including MovSeg~\cite{ji2026movseg}, Llama3-MS-CLIP~\cite{marimo2025llama3msclip}, and SATtxt~\cite{do2026sattxt}, explore vision-language learning with multispectral observations. Nevertheless, effectively exploiting complementary hyperspectral information for open-vocabulary pixel-level recognition without additional training remains challenging.

\section{Method}
\label{sec:method}

\begin{figure*}[ht]
    \centering
    \includegraphics[width=\textwidth]{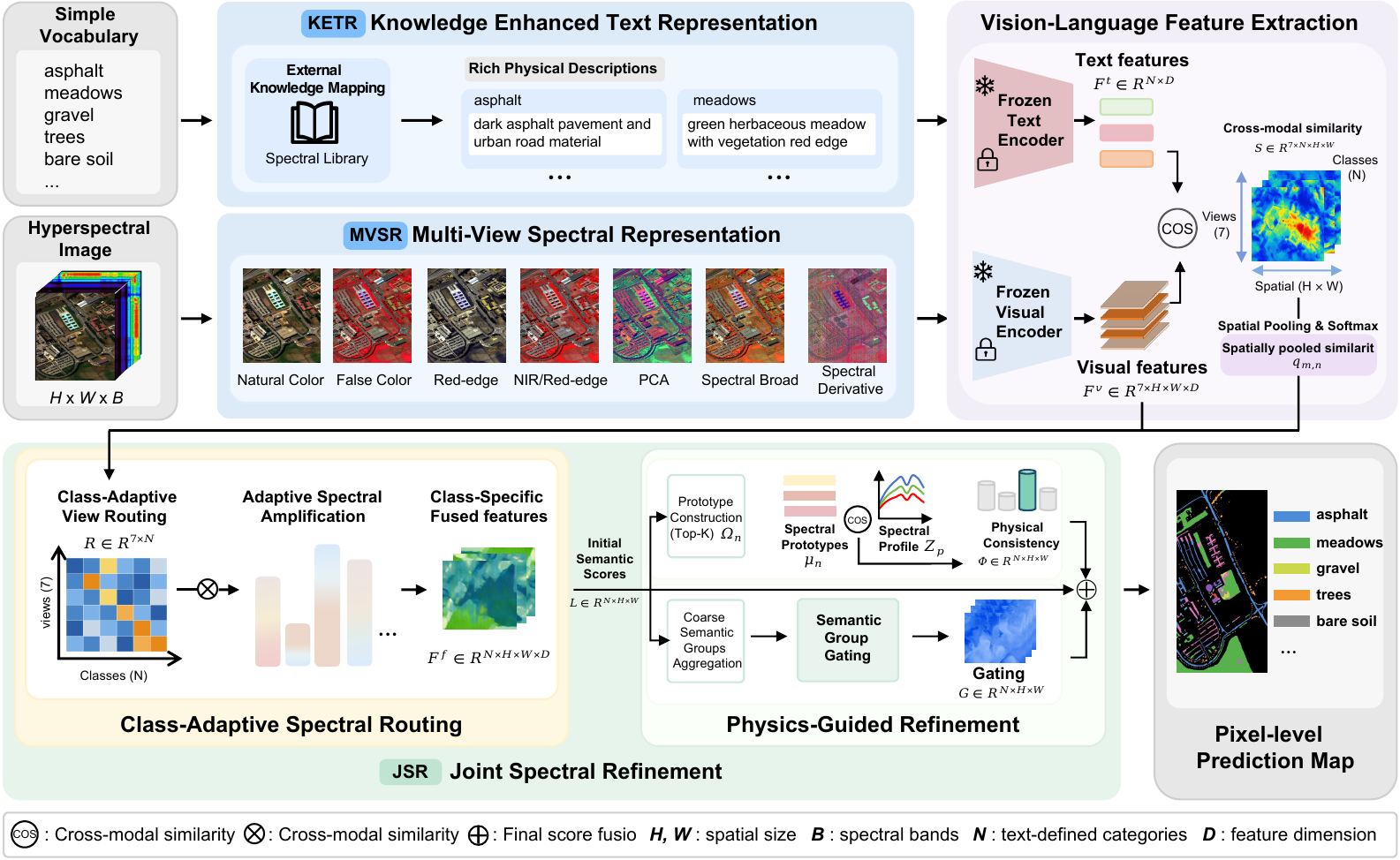}
    \vspace{-2em}
   \caption{Overview of SpectralRoute. The framework starts with knowledge-enhanced text representation and multiview spectral construction, followed by vision-language feature extraction using frozen CLIP to compute cross-modal similarities. Spatial pooling and softmax transform the similarities into routing weights. Class-adaptive spectral routing inputs both the routing weights and visual features to perform weighted summation and amplification. Physics-guided refinement integrates spectral prototypes, semantic gating, and initial scores to generate the final prediction.}
    \label{fig:framework}
    \vspace{-1.5em}
\end{figure*}

We propose \textbf{SpectralRoute}, a training-free framework for open-vocabulary hyperspectral semantic segmentation. As illustrated in Fig.~\ref{fig:framework}, SpectralRoute first constructs complementary spectral views and knowledge-enhanced text descriptions to form spectral-semantic associations via frozen CLIP encoders. It employs class-adaptive spectral routing to amplify category-relevant responses, and physics-guided refinement to cut fine-grained confusion.

\subsection{Problem Formulation}
\label{sec:problem}

Given a hyperspectral image $\mathbf{X}\in\mathbb{R}^{H\times W\times B}$, where $H$ and $W$ denote the spatial dimensions and $B$ is the number of spectral bands, we denote its wavelengths as $\boldsymbol{\Lambda}=\{\lambda_b\}_{b=1}^{B}$. The candidate vocabulary is defined as $\mathcal{C}=\{c_1,c_2,\ldots,c_N\}$, where $N$ denotes the number of text-defined categories. Our objective is to predict a pixel-level semantic map $\hat{\mathbf{Y}}\in\{1,\ldots,N\}^{H\times W}$:
\begin{equation}
\hat{\mathbf{Y}}=\mathcal{F}(\mathbf{X},\boldsymbol{\Lambda},\mathcal{C};\Theta_{\mathrm{frozen}}),
\label{eq:formulation}
\end{equation}
where $\mathcal{F}$ denotes the proposed framework and $\Theta_{\mathrm{frozen}}$ represents the pretrained vision-language model parameters. Both the visual encoder $\mathcal{E}_v$ and text encoder $\mathcal{E}_t$ remain frozen, allowing the vocabulary to change at inference time without target-domain annotations or parameter updates.

\subsection{Multi-View Spectral and Semantic Representations}
\label{sec:representation}

Pretrained VLMs accept three-channel inputs, making direct exploitation of high-dimensional hyperspectral information difficult. Compressing an HSI into a single visual representation may discard discriminative spectral cues, while simple names lack physical semantics. To address these limitations, we construct complementary spectral views and knowledge-enhanced text representations.

\noindent\textbf{Multi-View Spectral Representation.}
Different wavelength regions capture complementary properties of land-cover materials. We therefore transform the input HSI into $M=7$ three-channel views:
\begin{equation}
\mathcal{V}=\{\mathbf{V}_m\}_{m=1}^{M},\quad
\mathbf{V}_m=\Phi_m(\mathbf{X},\boldsymbol{\Lambda}),
\label{eq:multiview}
\end{equation}
where $\Phi_m$ denotes a predefined spectral transformation. The seven views are Natural Color, False Color, Red-edge, NIR/Red-edge, PCA, Spectral Broad, and Spectral Derivative. These views provide complementary information on visible appearance, vegetation conditions, and material reflectance. Notably, the Spectral Derivative view captures local spectral variations via first-order differentiation:
\begin{equation}
V_{\mathrm{grad}}(x,y)=
\frac{X_{\lambda+\Delta}(x,y)-X_{\lambda-\Delta}(x,y)}{2\Delta},
\label{eq:spectral_derivative}
\end{equation}
where $\Delta$ denotes the wavelength offset. Derivative responses at selected wavelengths are organized into a three-channel representation.

\noindent\textbf{Knowledge Enhanced Text Representation.}
While multiview representations preserve complementary spectral information, simple category names provide limited physical semantics. We therefore introduce an External Spectral Knowledge Mapping module that enriches each category $c_n$ with material properties and spectral characteristics from an external spectral library $\mathcal{K}_{\mathrm{spec}}$. The resulting description $d_n=\mathcal{G}(c_n,\mathcal{K}_{\mathrm{spec}})$ provides additional physical context without accessing target-domain annotations.

\noindent\textbf{Vision-Language Feature Extraction.}
The spectral views and enriched descriptions are fed into the frozen CLIP visual encoder $\mathcal{E}_v$ and text encoder $\mathcal{E}_t$, respectively. For each spectral view, the spatial visual features are obtained as:
\begin{equation}
\mathbf{F}_m^v=\mathcal{U}(\mathcal{E}_v(\mathbf{V}_m)),
\label{eq:visual_features}
\end{equation}
where $\mathcal{U}$ denotes spatial feature extraction and alignment. The text embeddings are given by $\mathbf{t}_n=\mathcal{E}_t(d_n)$. The resulting features are organized as $\mathbf{F}^v\in\mathbb{R}^{M\times H\times W\times D}$ and $\mathbf{F}^t\in\mathbb{R}^{N\times D}$, where $D$ is the shared feature dimension. We then compute the pixel-level cross-modal similarity:
\begin{equation}
S_{m,n}(p)=\frac{\mathbf{f}_m^v(p)^\top\mathbf{t}_n}{\|\mathbf{f}_m^v(p)\|_2\|\mathbf{t}_n\|_2},
\label{eq:cross_modal_similarity}
\end{equation}
where $p$ denotes a spatial position. The resulting tensor $\mathbf{S}\in\mathbb{R}^{M\times N\times H\times W}$ establishes semantic associations between spectral views and candidate categories. However, different views contribute unequally to individual categories, and uniform fusion may weaken discriminative spectral responses.

\subsection{Joint Spectral Refinement}
\label{sec:refinement}

\noindent\textbf{Class-Adaptive Spectral Routing.}
The discriminative importance of spectral views varies across land-cover categories. Uniform averaging or globally shared weights overlook these differences and may dilute informative responses. To address this limitation, we introduce spectral routing to dynamically evaluate view-category relevance and amplify informative spectral evidence.

\noindent\textbf{View-Class Evidence Evaluation.}
Given the semantic response tensor $\mathbf{S}$, we first compute the spatially pooled similarity $q_{m,n}=\frac{1}{HW}\sum_p S_{m,n}(p)$ for each view-category pair. Since semantic similarity alone may be affected by pretrained VLM biases, we further incorporate view-associated spectral evidence from the original HSI:
\begin{equation}
\bar{P}_{m,n}=\mathcal{Q}\left(q_{m,n},\Psi_m(\mathbf{X},\boldsymbol{\Lambda})\right),
\label{eq:spectral_evidence}
\end{equation}
where $\Psi_m$ extracts view-associated spectral information and $\mathcal{Q}$ denotes a parameter-free spectral evidence retrieval operation. The semantic response and spectral evidence are combined as $\hat{s}_{m,n}=\mathcal{A}(q_{m,n},\bar{P}_{m,n})$, where $\mathcal{A}$ denotes the evidence fusion operation.

\noindent\textbf{Class-Adaptive View Routing.}
Based on the joint relevance scores, we construct an image-conditioned routing matrix $\mathbf{R}\in\mathbb{R}^{M\times N}$ by normalizing across spectral views:
\begin{equation}
R_{m,n}=\frac{\exp(\hat{s}_{m,n}/\tau)}{\sum_{j=1}^{M}\exp(\hat{s}_{j,n}/\tau)},
\label{eq:adaptive_routing}
\end{equation}
where $\tau>0$ controls the concentration of routing weights. The matrix assigns category-specific importance to spectral views without introducing trainable parameters.

\noindent\textbf{Adaptive Spectral Amplification.}
To emphasize informative views while retaining complementary information, we construct category-conditioned visual features through parameter-free amplification:
\begin{equation}
\mathbf{F}_n^f(p)=\frac{1}{M+\gamma}\sum_{m=1}^{M}(1+\gamma R_{m,n})\mathbf{F}_m^v(p),
\label{eq:feature_fusion}
\end{equation}
where $\gamma\geq0$ controls the amplification strength. The normalization maintains a constant fusion weight, with $\gamma=0$ reducing to uniform averaging. The initial semantic score is computed as $L_n(p)=\langle\mathbf{F}_n^f(p),\mathbf{t}_n\rangle$. This mechanism enhances category-relevant spectral evidence, but semantic similarity alone may be insufficient to distinguish visually similar materials with different spectral properties.

\noindent\textbf{Physics-Guided Refinement.}
Despite class-adaptive routing, semantic predictions may confuse visually similar materials with distinct spectral properties, like natural grass and artificial turf. To address this limitation, we introduce physics-guided refinement combining spectral prototype consistency with hierarchical gating.

\noindent\textbf{Confidence-Guided Spectral Prototypes.}
Given the routed semantic scores $L_n(p)$, we obtain initial predictions $\hat{y}^{(0)}(p)=\arg\max_n L_n(p)$ and confidence scores $\pi_n(p)=\operatorname{Softmax}_n(L_n(p))$. For each category $n$, we select the top-$K$ confident pixels from its predicted region, denoted as $\Omega_n$. Using their original spectral profiles $\mathbf{z}_p\in\mathbb{R}^{B}$, we construct the category-specific prototype:
\begin{equation}
\boldsymbol{\mu}_n=\frac{1}{|\Omega_n|}\sum_{p\in\Omega_n}\mathbf{z}_p,
\label{eq:spectral_prototype}
\end{equation}
where $\Omega_n$ contains at most $K$ pixels and the prototype is computed only when $\Omega_n$ is nonempty. We then measure the spectral consistency between the spectral profile of each pixel and the corresponding category prototype:
\begin{equation}
\Phi(p,n)=\frac{\mathbf{z}_p^\top\boldsymbol{\mu}_n}{\|\mathbf{z}_p\|_2\|\boldsymbol{\mu}_n\|_2+\epsilon},
\label{eq:spectral_consistency}
\end{equation}
where $\epsilon$ ensures numerical stability. For categories without valid prototypes, we set $\Phi(p,n)=0$.

\noindent\textbf{Hierarchical Semantic Gating.}
Applying spectral correction to all categories may introduce irrelevant evidence. We organize the vocabulary into coarse semantic groups and compute a gating weight $G_n(p)$ by aggregating the prediction probabilities of categories within the group containing $n$. The final prediction score is defined as:
\begin{equation}
\tilde{L}_n(p)=L_n(p)+\lambda G_n(p)\Phi(p,n),
\label{eq:final_refinement}
\end{equation}
 where $\lambda$ controls the contribution of spectral consistency. This refinement incorporates spectral evidence while restricting its influence according to semantic relevance, reducing fine-grained category confusion.

\subsection{Training-Free Inference}
\label{sec:inference}

Given an input HSI and a candidate vocabulary, SpectralRoute constructs complementary spectral views and enriched text descriptions, extracts cross-modal responses using frozen CLIP encoders, and performs class-adaptive routing followed by physics-guided refinement. The final pixel-level segmentation map is obtained as:
\begin{equation}
\hat{Y}(p)=\arg\max_{n\in\{1,\ldots,N\}}\tilde{L}_n(p).
\label{eq:final_prediction}
\end{equation}

The process requires neither target-domain annotations nor parameter updates. When the vocabulary changes, the text embeddings and category-dependent responses can be recomputed without retraining, enabling training-free open-vocabulary hyperspectral segmentation.

\section{Experiments}
\begin{table*}[t]
\centering
\caption{Quantitative comparison of open-vocabulary semantic segmentation on three hyperspectral benchmarks. \protect\best{Best} and \protect\secondbest{second best} performances are highlighted. Our SpectralRoute achieves state-of-the-art results under the strict zero-shot setting.}
\label{tab:main_results}
\renewcommand{\arraystretch}{1.1}
\resizebox{\textwidth}{!}{
\begin{tabular}{l ccc ccc ccc}
\toprule
\multirow{2}{*}{\textbf{Methods}} & \multicolumn{3}{c}{\textbf{Pavia University}} & \multicolumn{3}{c}{\textbf{Washington DC Mall}} & \multicolumn{3}{c}{\textbf{Houston 2018}} \\
\cmidrule(lr){2-4} \cmidrule(lr){5-7} \cmidrule(lr){8-10}
 & OA (\%) & AA (\%) & Kappa (\%) & OA (\%) & AA (\%) & Kappa (\%) & OA (\%) & AA (\%) & Kappa (\%) \\
\midrule
\makebox[3.8cm][l]{SegEarth-OV~\cite{li2025segearth}\hfill{\scriptsize CVPR'25}} & 47.19 & 38.79 & 35.96 & 42.06 & 32.30 & 27.85 & 26.01 & \secondbest{27.60} & 17.92 \\
\makebox[3.8cm][l]{RSCLIP~\cite{wang2026rsclip}\hfill{\scriptsize TGRS'26}} & \secondbest{53.85} & 36.68 & 41.63 & 40.77 & 30.17 & 24.80 & 24.51 & 18.36 & 9.23 \\
\makebox[3.8cm][l]{RSKT-Seg~\cite{li2026rskt}\hfill{\scriptsize AAAI'26}} & 25.29 & 29.74 & 13.36 & 38.44 & 18.57 & 11.19 & \secondbest{35.79} & 23.02 & \secondbest{23.24} \\
\makebox[3.8cm][l]{HyperSIGMA~\cite{Wang_2025_HyperSIGMA}\hfill{\scriptsize TPAMI'25}} & 4.98 & 12.45 & 0.41 & 20.56 & 15.40 & 1.82 & 1.13 & 4.65 & -0.22 \\
\makebox[3.8cm][l]{SPECIAL~\cite{pang2026special}\hfill{\scriptsize TGRS'26}} & 52.13 & \secondbest{43.38} & \secondbest{41.70} & \secondbest{47.09} & \secondbest{38.47} & \secondbest{34.53} & 18.45 & 22.90 & 11.25 \\
\midrule
\makebox[3.8cm][l]{\textbf{SpectralRoute}\hfill{\scriptsize Ours}} & \best{70.10} & \best{49.38} & \best{58.21} & \best{54.96} & \best{41.50} & \best{43.60} & \best{40.99} & \best{47.51} & \best{32.23} \\
\bottomrule
\end{tabular}
}
\vspace{-1em}
\end{table*}

\begin{figure*}[t]
\centering
\captionsetup[subfigure]{font=footnotesize,justification=centering}

\begin{subfigure}[b]{0.13\textwidth}
    \centering
    \includegraphics[width=\linewidth]{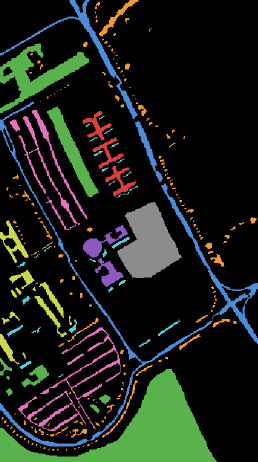}
    \caption{Ground Truth}
\end{subfigure}\hfill
\begin{subfigure}[b]{0.13\textwidth}
    \centering
    \includegraphics[width=\linewidth]{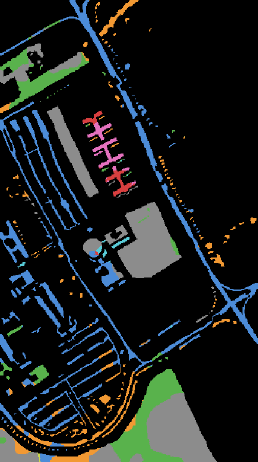}
    \caption{SegEarth-OV}
\end{subfigure}\hfill
\begin{subfigure}[b]{0.13\textwidth}
    \centering
    \includegraphics[width=\linewidth]{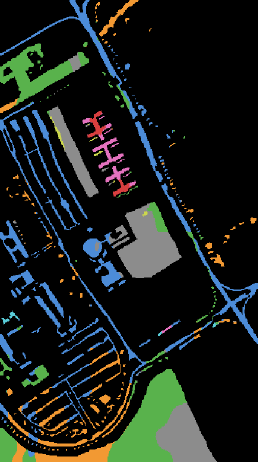}
    \caption{RSCLIP}
\end{subfigure}\hfill
\begin{subfigure}[b]{0.13\textwidth}
    \centering
    \includegraphics[width=\linewidth]{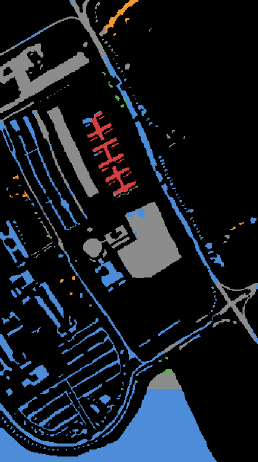}
    \caption{RSKT-Seg}
\end{subfigure}\hfill
\begin{subfigure}[b]{0.13\textwidth}
    \centering
    \includegraphics[width=\linewidth]{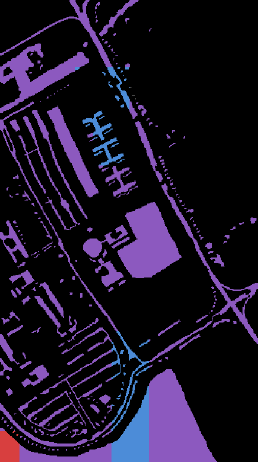}
    \caption{HyperSIGMA}
\end{subfigure}\hfill
\begin{subfigure}[b]{0.13\textwidth}
    \centering
    \includegraphics[width=\linewidth]{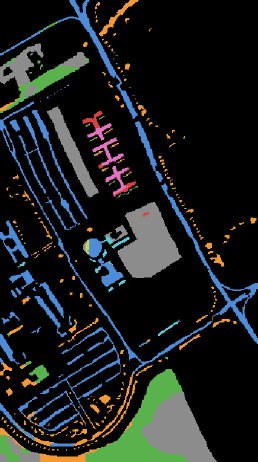}
    \caption{SPECIAL}
\end{subfigure}\hfill
\begin{subfigure}[b]{0.13\textwidth}
    \centering
    \includegraphics[width=\linewidth]{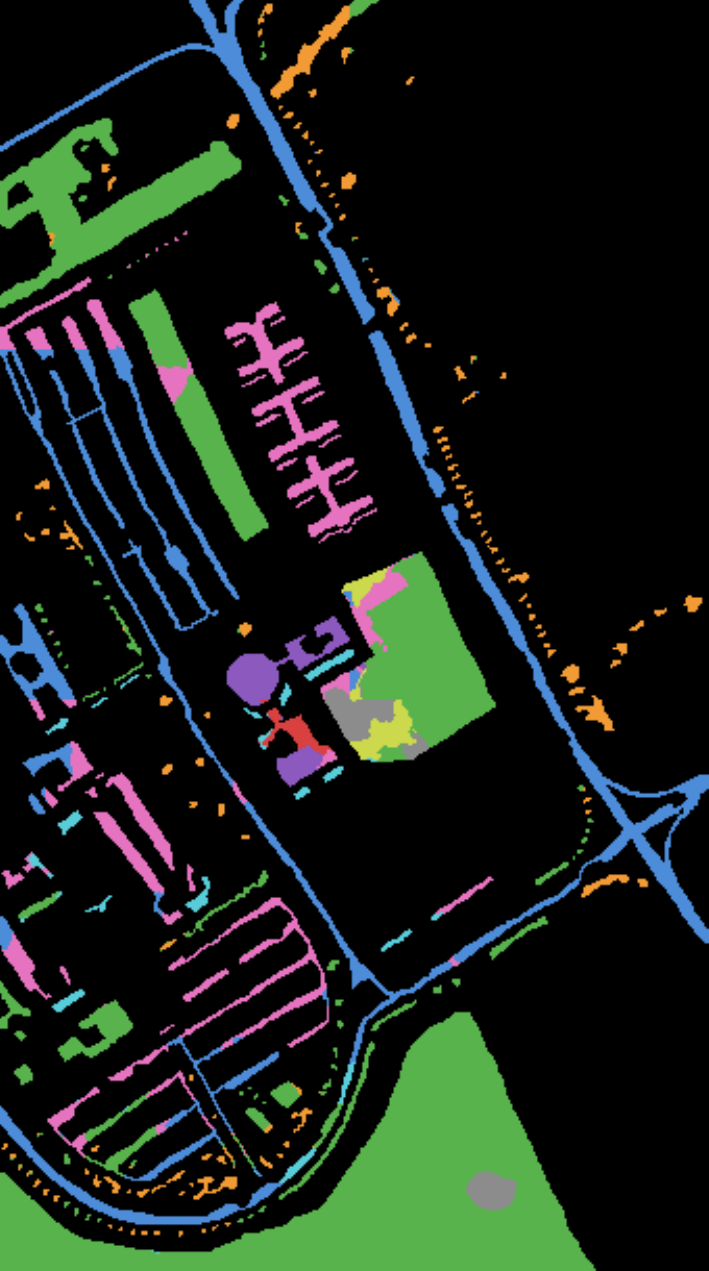}
    \caption{\textbf{Ours}}
\end{subfigure}
\vspace{1mm}
\begin{minipage}{0.95\textwidth}
    \centering
    \small 
    \renewcommand{\arraystretch}{0.9} 
    \textbf{Legend:}\hspace{0.6em}
    \begin{tabular}{lllll}
        \textcolor[RGB]{76,140,217}{\rule{1.5ex}{1.5ex}} Asphalt & 
        \textcolor[RGB]{89,179,77}{\rule{1.5ex}{1.5ex}} Meadows & 
        \textcolor[RGB]{204,217,77}{\rule{1.5ex}{1.5ex}} Gravel & 
        \textcolor[RGB]{242,154,51}{\rule{1.5ex}{1.5ex}} Trees & 
        \textcolor[RGB]{217,64,64}{\rule{1.5ex}{1.5ex}} Painted metal sheets \\
        \textcolor[RGB]{140,140,140}{\rule{1.5ex}{1.5ex}} Bare soil & 
        \textcolor[RGB]{140,89,191}{\rule{1.5ex}{1.5ex}} Bitumen & 
        \textcolor[RGB]{230,115,191}{\rule{1.5ex}{1.5ex}} Self-blocking bricks & 
        \textcolor[RGB]{89,204,217}{\rule{1.5ex}{1.5ex}} Shadows & 
        \textcolor[RGB]{0,0,0}{\rule{1.5ex}{1.5ex}} Unlabeled \\
    \end{tabular}
\end{minipage}
\vspace{-3mm}
\caption{Qualitative comparison of segmentation maps on the Pavia University dataset. Previous methods suffer from severe category confusion and fragmented boundaries, whereas SpectralRoute achieves sharper boundaries and finer category separation.}
\label{fig:vis}
\vspace{-2em}
\end{figure*}

\subsection{Experimental Setup}

\textbf{Datasets.} We evaluate SpectralRoute on three urban hyperspectral benchmarks: Pavia University~\cite{fauvel2008spectral}, Washington DC Mall~\cite{landgrebe2002hyperspectral}, and Houston 2018~\cite{le2018data}. These datasets cover different sensor characteristics and spatial resolutions. Houston 2018 is a highly challenging dataset with highly imbalanced categories and many spectrally similar materials, serving as a rigorous testbed for open-vocabulary recognition. For open-vocabulary evaluation, we adopt the standard class-name protocol without exposing any training images to the model. The
details of each dataset are provided in the Appendix \ref{sec:supp_dataset}.

\textbf{Implementation Details.} We employ a frozen CLIP with a ViT-B/16 visual backbone, without any target-domain fine-tuning. In the multi-view stage, we generate 7 complementary visual views. In the routing mechanism, we set the temperature parameter \(\tau\) to 0.1 and the scaling hyperparameter \(\gamma\) to 2.0, while the physical correction weight \(\lambda\) is set to 0.5. We implement the framework on a single NVIDIA RTX 4090 GPU. The entire inference operates in a purely training-free and zero-shot manner. See Appendix \ref{sec:supp_impl} for more details.

\textbf{Evaluation Metrics.} We adopt Overall Accuracy (OA), Average Accuracy (AA), and the Kappa coefficient. OA measures the global proportion of correctly classified pixels, but is highly sensitive to class imbalance. AA computes the mean accuracy across all categories, providing an unbiased evaluation of minority classes. Kappa measures the agreement between predictions and ground truth while accounting for random chance.

\subsection{Comparison with State-of-the-Art Methods}

We compare SpectralRoute with several state-of-the-art methods, including remote sensing OVSS models SegEarth-OV~\cite{li2025segearth}, RSCLIP~\cite{wang2026rsclip}, and RSKT-Seg~\cite{li2026rskt}, as well as hyperspectral foundation models HyperSIGMA~\cite{Wang_2025_HyperSIGMA} and SPECIAL~\cite{pang2026special}. To ensure a fair comparison, all open-vocabulary methods, including our SpectralRoute, are evaluated in a zero-shot manner without any target-domain fine-tuning. As shown in Table \ref{tab:main_results}, SpectralRoute outperforms all methods across the three benchmarks. The performance gap is most pronounced on the challenging Houston 2018 dataset, where SpectralRoute achieves 40.99\% OA and 47.51\% AA, surpassing the second-best methods—RSKT-Seg~\cite{li2026rskt} in OA and SegEarth-OV~\cite{li2025segearth} in AA—by 5.20 and 19.91 percentage points, respectively.The accuracy improvement empirically validates that our multi-view routing and physics-guided refinement effectively preserve spectral integrity and resolve visual ambiguities. Furthermore, our training-free method closely approaches fully-supervised closed-set models, demonstrating the immense potential of zero-shot segmentation for hyperspectral imagery.

We present qualitative visualizations for Pavia University in Fig. \ref{fig:vis} for intuitive assessment. Prior methods frequently confuse visually similar materials, whereas SpectralRoute generates more coherent, accurate segmentation outputs. The visual evidence demonstrates that dynamically routing category-specific spectral cues and leveraging physical priors resolves fine-grained ambiguities. More results can be found in Appendix \ref{sec:supp_results}.

\subsection{Ablation Study}

\begin{table*}[t]
\centering
\caption{Ablation study of the core modules across the three benchmarks. The baseline is progressively integrated with MVSR, KETR, and JSR to yield our full SpectralRoute.}
\label{tab:module}
\renewcommand{\arraystretch}{1.1}
\resizebox{\textwidth}{!}{
\begin{tabular}{ccc | ccc | ccc | ccc}
\toprule
\multicolumn{3}{c|}{\textbf{Modules}} & \multicolumn{3}{c|}{\textbf{Pavia University}} & \multicolumn{3}{c|}{\textbf{Washington DC Mall}} & \multicolumn{3}{c}{\textbf{Houston 2018}} \\
\cmidrule(lr){1-3} \cmidrule(lr){4-6} \cmidrule(lr){7-9} \cmidrule(lr){10-12}
\textbf{MVSR} & \textbf{KETR} & \textbf{JSR} & OA (\%) & AA (\%) & Kappa (\%) & OA (\%) & AA (\%) & Kappa (\%) & OA (\%) & AA (\%) & Kappa (\%) \\
\midrule
 & & & 50.42 & 41.84 & 39.98 & 35.78 & 29.73 & 22.56 & 14.51 & 26.26 & 10.58 \\
\checkmark & & & 56.56 & 42.06 & 46.03 & 45.41 & 36.54 & 32.56 & 12.20 & 23.75 & 9.57 \\
\checkmark & \checkmark & & 70.06 & 49.37 & 58.15 & 53.91 & 41.04 & 42.49 & 36.34 & 46.46 & 28.01 \\
\checkmark & \checkmark & \checkmark & \textbf{70.10} & \textbf{49.38} & \textbf{58.21} & \textbf{54.96} & \textbf{41.50} & \textbf{43.60} & \textbf{40.99} & \textbf{47.51} & \textbf{32.23} \\
\bottomrule
\end{tabular}
}
\end{table*}

\begin{figure*}[t]
    \centering
    \begin{subfigure}[b]{0.32\textwidth}
        \centering
        \includegraphics[width=\textwidth]{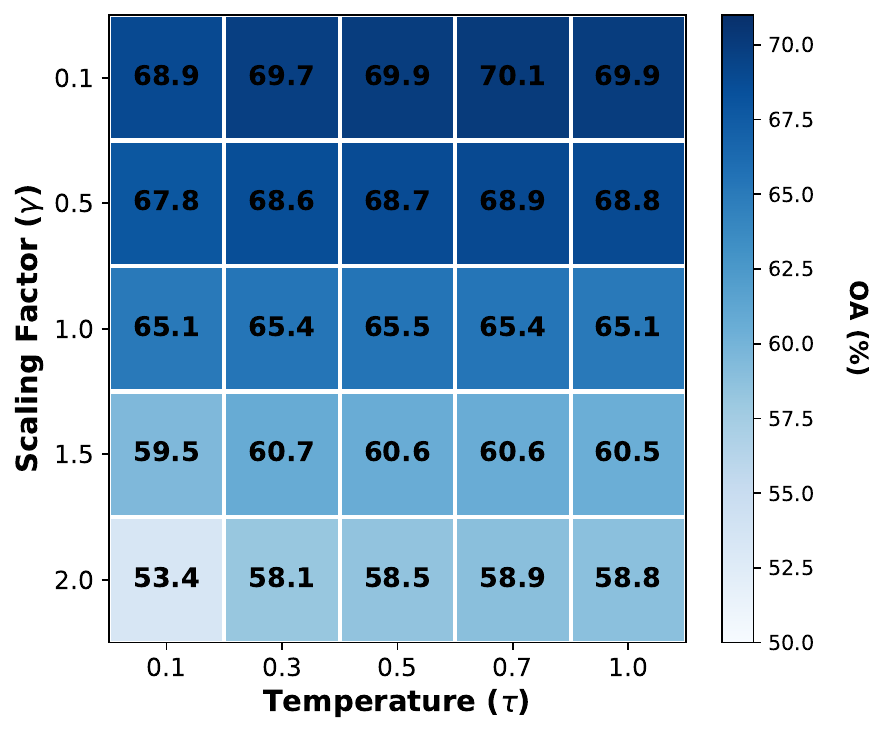}
        \caption{Pavia University}
    \end{subfigure}
    \hfill
    \begin{subfigure}[b]{0.32\textwidth}
        \centering
        \includegraphics[width=\textwidth]{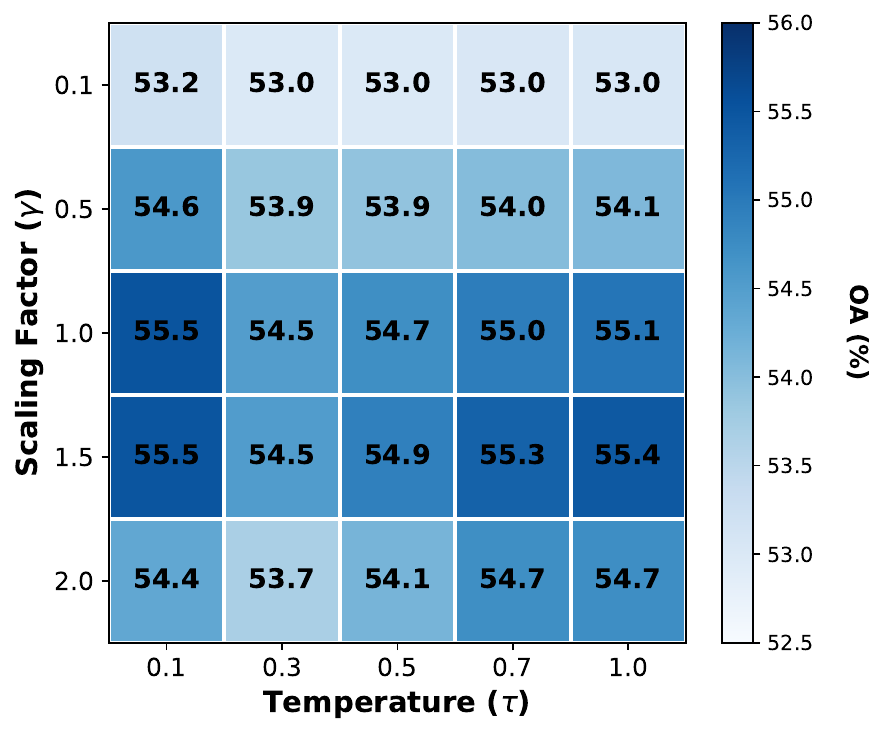}
        \caption{Washington DC Mall}
    \end{subfigure}
    \hfill
    \begin{subfigure}[b]{0.32\textwidth}
        \centering
        \includegraphics[width=\textwidth]{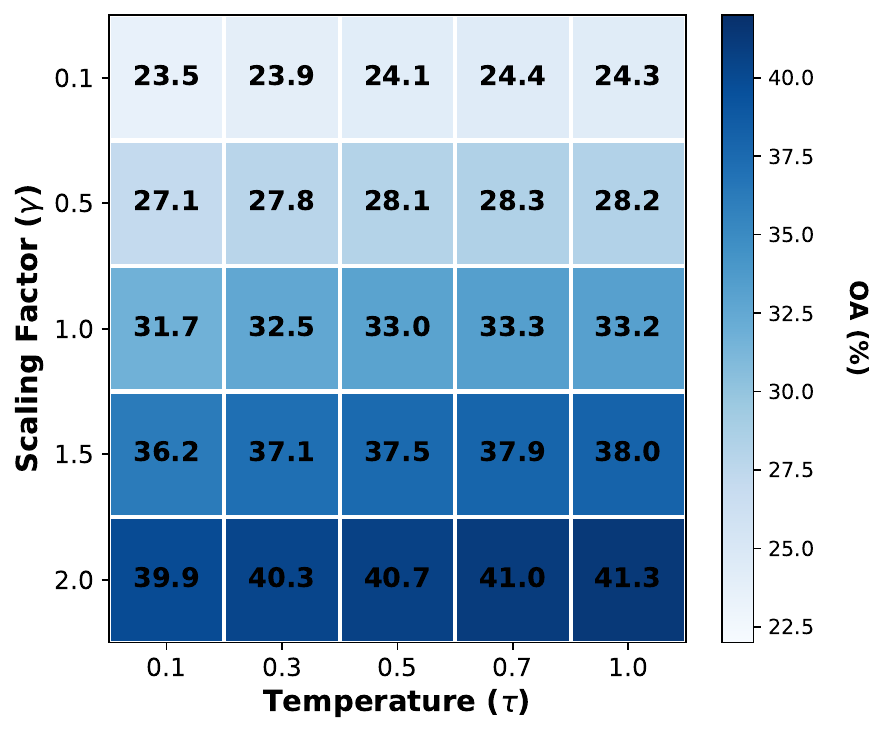}
        \caption{Houston 2018}
    \end{subfigure}
    \caption{Parameter sensitivity analysis of temperature \(\tau\) and scaling factor \(\gamma\). While simpler scenes favor smaller values, highly complex scenes rely more on strong spectral amplification. Our method adaptively scales with scene complexity and maintains robust performance.}
    \label{fig:sensitivity}
    \vspace{-1em}
\end{figure*}

\textbf{Effectiveness of Core Modules.} Table \ref{tab:module} reveals distinct physical behaviors across datasets. On the simpler Pavia University, adding Multi-View Spectral Representation (MVSR) and Knowledge Enhanced Text Representation (KETR) injects physical and semantic priors, yielding semantic saturation. In contrast, the complex Houston 2018 dataset exposes the limitations of uniform fusion. Here, naively adding MVSR causes a performance drop due to spectral redundancy. However, KETR acts as a semantic anchor, resolving the conflict. More importantly, incorporating our Joint Spectral Refinement (JSR) module yields substantial gains, pushing the Houston OA to 40.99\% and AA to 47.51\%. The ablation results demonstrate that while simple scenes benefit from multimodal alignment, complex scenes rely on our joint routing and Physics-Guided Refinement (PGR) mechanisms.

\textbf{Comparison with Fusion Strategies.} Table \ref{tab:fusion} compares our Joint Spectral Refinement module with conventional fusion strategies on Washington DC Mall. Traditional methods rely on globally shared weights or max-confidence, failing to resolve inter-category conflicts. While traditional methods achieve a comparable OA around 51\%, the corresponding AA drops drastically to around 37\%, indicating a severe bias toward majority classes. In contrast, our module dynamically assigns category-specific weights, significantly improving the OA to 54.96\%, AA to 41.50\%, and Kappa to 43.60\%.

\begin{table}[t]
\centering
\caption{Comparison against conventional fusion strategies on the Washington DC Mall dataset.}
\label{tab:fusion}
\renewcommand{\arraystretch}{1.1}
\small
\begin{tabular*}{\columnwidth}{@{\hspace{0.8em}\extracolsep{\fill}} lccc @{\hspace{0.8em}}}
\toprule
\textbf{Fusion Strategy} & \textbf{OA (\%)} & \textbf{AA (\%)} & \textbf{Kappa (\%)} \\
\midrule
Uniform Average & 51.62 & 38.65 & 39.57 \\
Max Confidence & 50.72 & 36.66 & 38.05 \\
Ours (JSR) & \textbf{54.96} & \textbf{41.50} & \textbf{43.60} \\
\bottomrule
\end{tabular*}
\end{table}

\textbf{Effect of Physics-Guided Refinement.} Table \ref{tab:houston_case} provides a fine-grained analysis of three visually similar categories—grass, buildings, and parking lots—on the challenging Houston 2018 dataset. Without physical constraints (w/o PGR), the model struggles to distinguish these categories due to similar RGB appearances. With PGR, representative examples show clear gains: Artificial Turf rises from 0.00\% to 45.36\%, Non-residential Buildings from 12.09\% to 48.48\%, and Paved Parking Lots from 3.30\% to 15.46\%. Such refinement aligns with our motivation: physical priors disentangle intrinsic material differences.

\begin{table}[t]
\centering
\caption{Class-wise accuracy (\%) for visually similar land-cover classes on Houston 2018, without and with PGR.}
\label{tab:houston_case}
\small
\renewcommand{\arraystretch}{1.1}
\begin{tabular*}{\columnwidth}{@{\hspace{0.8em}\extracolsep{\fill}} lcc @{\hspace{0.8em}}}
\toprule
\textbf{Class} & \textbf{w/o PGR} & \textbf{w/ PGR} \\
\midrule
Healthy Grass          & 99.62 & 97.87\ \textcolor{red!80!black}{\scriptsize($-$1.75\,$\downarrow$)} \\
Stressed Grass         & 5.50  & 22.96\ \textcolor{green!40!black}{\scriptsize(+17.46\,$\uparrow$)} \\
Artificial Turf        & 0.00  & 45.36\ \textcolor{green!40!black}{\scriptsize(+45.36\,$\uparrow$)} \\
Residential Buildings  & 28.89 & 55.19\ \textcolor{green!40!black}{\scriptsize(+26.30\,$\uparrow$)} \\
Non-Res.\ Buildings    & 12.09 & 48.48\ \textcolor{green!40!black}{\scriptsize(+36.39\,$\uparrow$)} \\
Paved Parking Lots     & 3.30  & 15.46\ \textcolor{green!40!black}{\scriptsize(+12.16\,$\uparrow$)} \\
Unpaved Parking Lots   & 76.15 & 97.44\ \textcolor{green!40!black}{\scriptsize(+21.29\,$\uparrow$)} \\
\bottomrule
\end{tabular*}
\end{table}

\subsection{Parameter Sensitivity and Efficiency}

\textbf{Sensitivity to Hyperparameters.} We evaluate the sensitivity of \(\tau\) and \(\gamma\). As shown in Fig. \ref{fig:sensitivity}, optimal hyperparameters exhibit a strong dependence on scene complexity. The simple Pavia University prefers lower values, while the highly complex Houston 2018 requires larger values to resolve severe spectral ambiguities. The parameter behavior confirms that our method adaptively scales with scene complexity and maintains robust performance.

\textbf{Inference Efficiency.} As shown in Table \ref{tab:efficiency}, the peak memory of SpectralRoute is significantly lower than that of SegEarth-OV and SPECIAL, and is comparable to HyperSIGMA and RSCLIP. Notably, SegEarth-OV and SPECIAL exceed the 24 GB memory limit under default configurations; we adapt their running configurations without altering model architectures. Although SpectralRoute takes 52.13s compared to the ~12s of single-view methods, the latency is primarily because we must sequentially process 7 distinct visual views. However, instead of naively loading the model 7 times, we share the frozen visual backbone and text encoder across all views with a single model load. The key trade-off lies in exchanging a moderate latency overhead for a 5.4\(\times\) reduction in memory footprint and significant accuracy gains. All evaluated methods operate under a training-free paradigm.

\begin{table}[t]
\centering
\caption{Inference efficiency comparison on the Pavia University.}
\label{tab:efficiency}
\small
\renewcommand{\arraystretch}{1.1}
\begin{tabular*}{1.0\columnwidth}{@{\hspace{0.8em}\extracolsep{\fill}} lcc @{\hspace{0.8em}}}
\toprule
\textbf{Method} & \textbf{Time (s)} & \textbf{Memory (GB)} \\
\midrule
\makebox[3.6cm][l]{SegEarth-OV~\cite{li2025segearth}\hfill{\scriptsize CVPR'25}} & 14.46 & 6.22 \\
\makebox[3.6cm][l]{RSCLIP~\cite{wang2026rsclip}\hfill{\scriptsize TGRS'26}} & 12.80 & 1.12 \\
\makebox[3.6cm][l]{RSKT-Seg~\cite{li2026rskt}\hfill{\scriptsize AAAI'26}} & 11.86 & 2.10 \\
\makebox[3.6cm][l]{HyperSIGMA~\cite{Wang_2025_HyperSIGMA}\hfill{\scriptsize TPAMI'25}} & 12.69 & 1.04 \\
\makebox[3.6cm][l]{SPECIAL~\cite{pang2026special}\hfill{\scriptsize TGRS'26}} & 27.36 & 6.24 \\
\makebox[3.6cm][l]{\textbf{SpectralRoute}\hfill{\scriptsize Ours}} & 52.13 & 1.15 \\
\bottomrule
\end{tabular*}
\end{table}

\subsection{Spectral Contribution Analysis}

\begin{figure}[t]
    \centering
    \includegraphics[width=\columnwidth]{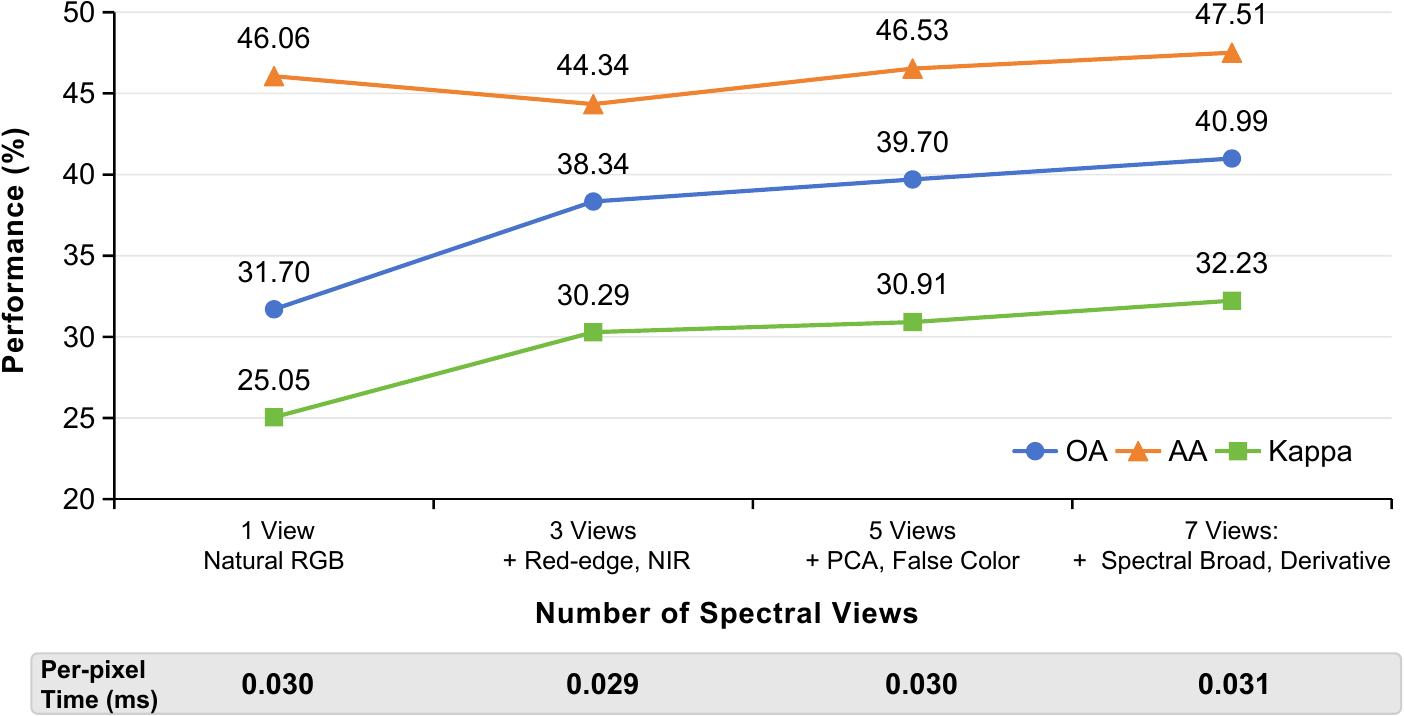}
    \caption{Effect of progressively adding spectral views from spatial appearance to material variations on the Houston 2018. Per-pixel time is measured after amortizing fixed overheads.}
    \label{fig:views}
    \vspace{-1em}
\end{figure}

\textbf{Analysis of Spectral Contributions.} We investigate how different spectral representations contribute to the final prediction by progressively adding views along a physically motivated cognitive chain from spatial appearance to material variations. As shown in Fig. \ref{fig:views}, the single Natural RGB view achieves an OA of 31.70\% but suffers from a biased AA of 46.06\%, indicating that relying solely on spatial appearance fails to recognize complex categories. Introducing vegetation-sensitive Red-edge and NIR views significantly improves the OA to 38.34\%. Incorporating data-driven PCA and False Color views further boosts the OA to 39.70\%. Notably, the addition of material-level views, including Spectral Broad and Spectral Derivative, further elevates the OA to 40.99\% and AA to 47.51\%. The ablation trend demonstrates that while the first five views capture dominant patterns, the material-level views act as a crucial side-channel for correcting fine-grained misclassifications.

\textbf{Computational Efficiency.} The per-pixel inference time remains highly stable at approximately 0.030 ms across all view configurations. The negligible variation reveals a clear amortization effect: the fixed computational overheads, including file I/O and model initialization, are effectively amortized by the dense pixel operations. As the number of views increases, the expanded feature matrices further dilute these fixed overheads, resulting in a nearly constant per-pixel latency. Therefore, our multi-view mechanism introduces negligible additional computational burden, ensuring high practical feasibility.

\section{Conclusion}
In this paper, we address the challenging task of open-vocabulary hyperspectral semantic segmentation. We highlight that directly applying RGB-pretrained vision-language models to HSIs is fundamentally limited by the 3-channel input constraint, while conventional uniform multiview fusion dilutes category-specific spectral evidence. To overcome these limitations, we propose SpectralRoute, a dedicated training-free framework. By decomposing HSIs into multiple complementary physical views and enriching text descriptions with external spectral knowledge, our method successfully bridges the modality gap. Furthermore, the proposed class-adaptive spectral routing dynamically amplifies the most informative spectral evidence for each category, and the physics-guided refinement effectively resolves ambiguities between visually similar materials. Extensive experiments on three urban hyperspectral benchmarks demonstrate that SpectralRoute consistently outperforms state-of-the-art methods in a strictly zero-shot manner, while maintaining a remarkably low memory footprint.

Despite these advancements, accurately distinguishing fine-grained categories with highly similar spectral responses remains an open challenge. Future work will investigate sensor-agnostic spectral-language pre-training, uncertainty-aware routing for overlapping spectral signatures, and physically grounded cross-sensor adaptation to establish more reliable associations between continuous spectral characteristics and natural language concepts.

\section*{LLM Usage Statement}


Large language models (LLMs) were used in this study for two purposes. First, they assist in grammar checking and manuscript polishing to improve readability and academic phrasing. Second, within the Spectral Knowledge-Enhanced Semantics (KETR) module, LLMs generate enriched class descriptions grounded in external spectral knowledge for spectral-semantic alignment.

No LLMs were used to generate core scientific ideas, design the methodology, or create the main experimental results. All research contributions, algorithmic designs, and data visualizations were entirely completed by the authors. The authors reviewed and verified all AI-assisted content and take full responsibility for the accuracy, integrity, and originality of the entire work.
\bibliographystyle{ieeenat_fullname}
\bibliography{main}

\clearpage
\setcounter{page}{1}
\maketitlesupplementary
\appendix
\setcounter{section}{0}
\renewcommand{\thesection}{\Alph{section}}

The ability to recognize land-cover materials beyond predefined categories is increasingly important for scalable Earth observation, as practical mapping applications must accommodate evolving recognition requirements, diverse sensing platforms, and substantial variations across geographic environments. Hyperspectral imagery (HSI) provides rich spectral information for distinguishing materials with similar visual appearances, yet conventional segmentation methods often rely on extensive annotations and task-specific adaptation. Bridging continuous spectral observations with language-driven recognition therefore offers a promising direction toward more flexible and accessible HSI interpretation. To this end, we propose \textbf{SpectralRoute}, a training-free framework that integrates Multi-View Spectral Representation (MVSR), Knowledge-Enhanced Text Representation (KETR), and Joint Spectral Refinement (JSR). By combining complementary spectral evidence, knowledge-enhanced semantics, and physics-guided refinement, SpectralRoute advances the use of frozen vision-language models for fine-grained open-vocabulary HSI segmentation, offering a promising foundation for scalable and adaptable Earth observation applications.

The supplementary material provides additional implementation details and experimental evidence to support the main paper. Sec.~\ref{sec:supp_dataset} describes the characteristics of the three HSI benchmarks, followed by implementation details and experimental settings in Sec.~\ref{sec:supp_impl}. Sec.~\ref{sec:supp_results} presents additional quantitative and qualitative results, including class-wise accuracy and segmentation visualizations. Sec.~\ref{sec:supp_analysis} provides extended ablation studies and analyses of spectral views, fusion strategies, and physics-guided refinement. Finally, Sec.~\ref{sec:supp_future} discusses future research directions.

\section{Dataset Description}
\label{sec:supp_dataset}

Table~\ref{tab:supp_datasets} summarizes the acquisition characteristics and annotation settings of the three urban hyperspectral benchmarks. Pavia University was acquired over Pavia, Italy, by the ROSIS sensor~\cite{fauvel2008spectral}. Washington DC Mall was captured over the National Mall area in Washington, DC, by the HYDICE sensor~\cite{landgrebe2002hyperspectral}. Houston 2018 covers the University of Houston campus and its surrounding area and was collected by the CASI sensor~\cite{le2018data}. The three benchmarks differ in sensor properties, spatial sampling, scene extent, and class composition, enabling evaluation under complementary urban spectral conditions.

\begin{table}[H]
\centering
\caption{Overview of the three hyperspectral benchmarks, including sensor configurations, spatial dimensions, spectral bands, spatial resolutions, and class numbers.}
\label{tab:supp_datasets}
\footnotesize
\renewcommand{\arraystretch}{1.1}
\setlength{\tabcolsep}{4pt}
\resizebox{\columnwidth}{!}{
\begin{tabular}{lccccc}
\toprule
\textbf{Dataset} & \textbf{Sensor} & \textbf{Size} & \textbf{Bands} & \textbf{Resolution} & \textbf{Classes} \\
\midrule
Pavia University~\cite{fauvel2008spectral} & ROSIS & $610 \times 340$ & 103 & 1.3 m & 9 \\
Washington DC Mall~\cite{landgrebe2002hyperspectral} & HYDICE & $1280 \times 307$ & 191 & 2.8 m & 7 \\
Houston 2018~\cite{le2018data} & CASI & $601 \times 2384$ & 50 & 1.0 m & 20 \\
\bottomrule
\end{tabular}
}
\end{table}

Pavia University contains compact urban structures with clear spatial boundaries, including roads, buildings, vegetation, and impervious surfaces. Washington DC Mall covers a larger urban region with substantial spatial variation and mixed land-cover patterns. Houston 2018 presents the most challenging setting because the benchmark contains highly imbalanced categories and multiple materials with similar visual appearances but distinct spectral responses. The experiments use the standard class names provided by each benchmark as the candidate vocabulary. SpectralRoute receives neither target-domain training images nor pixel-level annotations during inference.

\section{Implementation Details}
\label{sec:supp_impl}

\subsection{Multi-View Spectral Construction}
\label{sec:supp_mvsr}

SpectralRoute constructs seven complementary three-channel views from each input hyperspectral image: Natural Color, False Color, Red-edge, Near-Infrared (NIR)/Red-edge, Principal Component Analysis (PCA), Spectral Broad, and Spectral Derivative. Each view is generated through a predefined spectral transformation, providing a distinct representation of the original spectral measurements. Together, the seven views capture complementary cues from visible appearance, vegetation-related responses, and material-dependent spectral variations. All transformations are deterministic, contain no learnable parameters, and require no target-domain supervision.

Natural Color maps representative visible blue, green, and red bands to the three channels, preserving the spatial structures and color patterns of dominant urban objects. False Color replaces the visible red channel with a near-infrared response, enhancing the contrast between vegetation and artificial surfaces. Red-edge combines red-edge and visible spectral responses to highlight vegetation-related reflectance transitions, while NIR/Red-edge integrates near-infrared and red-edge information to strengthen the distinction among vegetation, water, and non-vegetated surfaces. PCA projects the full hyperspectral cube onto its first three principal components, retaining dominant spectral variations while reducing inter-band redundancy. Spectral Broad combines material-sensitive spectral regions to enhance reflectance and absorption differences among soil, gravel, pavement, and built surfaces. Spectral Derivative computes first-order differences between adjacent spectral bands and maps selected derivative responses to three channels, highlighting local spectral slopes, absorption valleys, and reflectance peaks.

The Natural Color view uses dataset-specific Red, Green, and Blue (RGB) band triplets of $(52, 28, 5)$ for Pavia University, $(49, 35, 14)$ for Washington DC Mall, and $(20, 12, 5)$ for Houston 2018. These triplets are selected independently to account for the different wavelength samplings of the ROSIS, HYDICE, and CASI sensors. For each dataset, the selected bands correspond to the available wavelengths closest to the target red, green, and blue regions, respectively. The remaining views use sensor-appropriate spectral bands according to their respective construction strategies. In particular, Red-edge and NIR/Red-edge select bands from their corresponding wavelength regions, approximately 690--730 nm and 750--900 nm, respectively. Spectral Broad uses three predefined band indices for each dataset to construct a representation spanning a broader portion of the available spectral range. For Spectral Derivative, the three channels are associated with target wavelengths of approximately 450, 550, and 680 nm. The corresponding band indices are dynamically determined for each dataset by selecting the available wavelengths closest to these targets, rather than using fixed indices across datasets. Each three-channel view is independently normalized before feature extraction to maintain a consistent input range across the seven spectral transformations.

\subsection{Knowledge-Enhanced Text Representation}
\label{sec:supp_text}

Open-vocabulary hyperspectral segmentation typically adopts short category names as text queries. A bare category label often lacks adequate information to characterize the material traits and spectral signatures that separate different hyperspectral classes. This limitation stands out with frozen VLMs, whose text encoders receive no adaptation toward the hyperspectral target domain. SpectralRoute adopts an offline knowledge-enhanced vocabulary expansion module to acquire richer and more discriminative semantic cues under strict training-free inference.

We build a global class description repository covering all land-cover categories from the three datasets. Each category holds two forms of complementary information: alternative aliases, alongside descriptive records encoding semantic and physical properties of the underlying ground material. While aliases broaden lexical coverage, descriptions provide additional context about material composition, visual appearance, structural properties, and characteristic spectral responses. These descriptions are initially curated based on dataset category names and remote sensing domain knowledge, with assistance from large language models (LLMs) for further expansion and refinement. The resulting repository is organized as a global vocabulary, allowing categories shared across different datasets to reuse the same descriptions. To maintain a consistent and minimal vocabulary interface, every dataset comes with a lightweight vocabulary definition, storing category indices, names, and raw text prompts. Relying on the built-in vocabulary expansion unit, the framework matches incoming category names against repository entries through case-insensitive exact matching, retrieves the associated aliases and descriptions, and incorporates them alongside the original queries. If no matching entry is found, the original category query is retained without additional expansion. Both aliases and descriptions are retained as auxiliary fields inside the expanded vocabulary.

The original category queries, together with all associated aliases and physical descriptions, are individually fed into the frozen text encoder to generate text embeddings. All resulting embeddings participate in feature aggregation and are averaged across the text inputs to obtain a single 512-dimensional representation for each category. Drawing on diverse lexical expressions and physically grounded descriptions, the resulting representations provide enriched semantic signals for cross-modal matching against multi-view hyperspectral features. Since the knowledge repository is prepared offline, this process requires neither online knowledge retrieval nor on-the-fly text generation during inference, and introduces no additional model fine-tuning.

\begin{table*}[t]
\centering
\caption{Class-wise accuracy comparison of open-vocabulary semantic segmentation methods on the Pavia University dataset. \protect\best{Best} and \protect\secondbest{second best} performances are highlighted.}
\label{tab:supp_pavia_perclass}
\small
\renewcommand{\arraystretch}{1.1}
\setlength{\tabcolsep}{4pt}
\begin{tabular*}{\textwidth}{@{\hspace{0.8em}\extracolsep{\fill}}lcccccc@{\hspace{0.8em}}}
\toprule
\textbf{Class}
& \textbf{SegEarth-OV~\cite{li2025segearth}}
& \textbf{RSCLIP~\cite{wang2026rsclip}}
& \textbf{RSKT-Seg~\cite{li2026rskt}}
& \textbf{HyperSIGMA~\cite{Wang_2025_HyperSIGMA}}
& \textbf{SPECIAL~\cite{pang2026special}}
& \textbf{SpectralRoute} \\
\midrule
Asphalt
& 87.35 & \secondbest{82.10} & 62.86 & 12.08 & \secondbest{90.47} & \best{93.74} \\

Meadows
& 35.81 & \secondbest{57.23} & 0.39 & 0.00 & 41.01 & \best{96.99} \\

Gravel
& 0.00 & 0.00 & 0.00 & 0.00 & 0.00 & 0.00 \\

Trees
& 68.02 & 60.41 & 16.61 & 0.00 & \best{95.53} & \secondbest{64.65} \\

Painted Metal Sheets
& \secondbest{42.60} & 38.14 & \best{91.67} & 0.00 & 29.00 & 0.00 \\

Bare Soil
& \secondbest{97.16} & 90.20 & 96.12 & 0.00 & \best{99.46} & 11.89 \\

Bitumen
& 0.00 & 0.00 & 0.00 & \best{100.00} & 0.00 & \secondbest{76.47} \\

Self-Blocking Bricks
& 0.00 & 0.00 & 0.00 & 0.00 & 0.00 & \best{41.50} \\

Shadows
& 18.16 & 2.01 & 0.00 & 0.00 & \secondbest{34.95} & \best{59.13} \\
\midrule
OA (\%)
& 47.19 & \secondbest{53.85} & 25.29 & 4.98 & 52.13 & \best{70.10} \\

AA (\%)
& 38.79 & 36.68 & 29.74 & 12.45 & \secondbest{43.38} & \best{49.38} \\

Kappa (\%)
& 35.96 & 41.63 & 13.36 & 0.41 & \secondbest{41.70} & \best{58.21} \\
\bottomrule
\end{tabular*}
\end{table*}

\subsection{Routing and Physics-Guided Refinement}
\label{sec:supp_routing}

\textbf{Class-Adaptive Spectral Routing.} Different land-cover categories rely on distinct spectral cues for discrimination. A uniform fusion strategy may therefore overlook category-specific spectral characteristics. To address this limitation, SpectralRoute introduces Class-Adaptive Spectral Routing, which estimates the relevance of each spectral view to individual categories without introducing additional trainable parameters. The cross-modal similarity responses are first spatially aggregated to characterize view-category associations. To reduce the influence of potential biases in pretrained VLMs, spectral evidence is also retrieved from the original hyperspectral image. Combining semantic relevance with view-associated spectral evidence forms the basis for category-specific routing.

Such joint relevance scores are normalized across spectral views using softmax to generate a category-specific routing matrix. These routing weights guide multi-view feature fusion, assigning greater contributions to views that provide more relevant information for each category while retaining complementary spectral cues. An adaptive amplification mechanism further enhances category-relevant spectral responses. Unlike uniform fusion, this strategy allows the fused representations to reflect the distinct spectral characteristics of individual land-cover categories.

\textbf{Physics-Guided Refinement.} Although category-adaptive routing improves spectral feature representation, semantic ambiguity may persist for visually similar yet spectrally distinct materials. To address this issue, SpectralRoute introduces Physics-Guided Refinement (PGR) to incorporate spectral consistency into the initial semantic predictions. For each category, the top-$K$ most confident pixels are selected to construct a category-specific spectral prototype from their original hyperspectral profiles, with $K=100$ for all three datasets. The spectral consistency between individual pixels and these prototypes provides additional physical evidence for category discrimination. If no valid prototype is available for a category, its spectral consistency response is set to zero. 

Hierarchical semantic gating further regulates the influence of spectral correction by aggregating prediction probabilities within predefined coarse semantic groups. These groups are configured according to the category composition of each dataset. The gated spectral consistency is then integrated with the initial semantic scores, allowing physical evidence to refine predictions while restricting corrections to semantically related categories.

\section{Additional Experimental Results}
\label{sec:supp_results}

\subsection{Class-Wise Accuracy}
\label{sec:supp_classwise}

\begin{table*}[t]
\centering
\caption{Class-wise accuracy comparison of open-vocabulary semantic segmentation methods on the Washington DC Mall dataset. \protect\best{Best} and \protect\secondbest{second best} performances are highlighted.}
\label{tab:supp_washington_perclass}
\small
\renewcommand{\arraystretch}{1.1}
\setlength{\tabcolsep}{5pt}
\begin{tabular*}{\textwidth}{@{\hspace{0.8em}\extracolsep{\fill}}lcccccc@{\hspace{0.8em}}}
\toprule
\textbf{Class}
& \textbf{SegEarth-OV~\cite{li2025segearth}}
& \textbf{RSCLIP~\cite{wang2026rsclip}}
& \textbf{RSKT-Seg~\cite{li2026rskt}}
& \textbf{HyperSIGMA~\cite{Wang_2025_HyperSIGMA}}
& \textbf{SPECIAL~\cite{pang2026special}}
& \textbf{SpectralRoute} \\
\midrule
Shadow
& 0.00 & 0.00 & 0.00 & 0.00 & 0.00 & \best{4.09} \\

Roof
& 26.22 & 8.68 & 0.00 & 0.00 & \secondbest{33.52} & \best{53.65} \\

Water
& 38.84 & \best{53.19} & 0.00 & 0.00 & \secondbest{53.01} & 32.75 \\

Street
& 72.24 & \secondbest{78.73} & 42.73 & \best{96.19} & 74.47 & 78.98 \\

Grass
& 30.25 & \secondbest{43.57} & \best{85.77} & 0.00 & 29.68 & \secondbest{45.19} \\

Path
& 0.36 & 1.26 & 0.00 & 0.00 & \best{6.85} & \secondbest{4.02} \\

Trees
& 58.21 & 25.79 & 1.51 & 11.62 & \secondbest{71.74} & \best{71.79} \\
\midrule
OA (\%)
& 42.06 & 40.77 & 38.44 & 20.56 & \secondbest{47.09} & \best{54.96} \\

AA (\%)
& 32.30 & 30.17 & 18.57 & 15.40 & \secondbest{38.47} & \best{41.50} \\

Kappa (\%)
& 27.85 & 24.80 & 11.19 & 1.82 & \secondbest{34.53} & \best{43.60} \\
\bottomrule
\end{tabular*}
\end{table*}

\begin{figure*}[t]
    \centering
    \captionsetup[subfigure]{font=footnotesize,justification=centering}

    \begin{subfigure}[b]{0.135\textwidth}
        \centering
        \includegraphics[width=\linewidth,height=0.25\textheight,keepaspectratio]{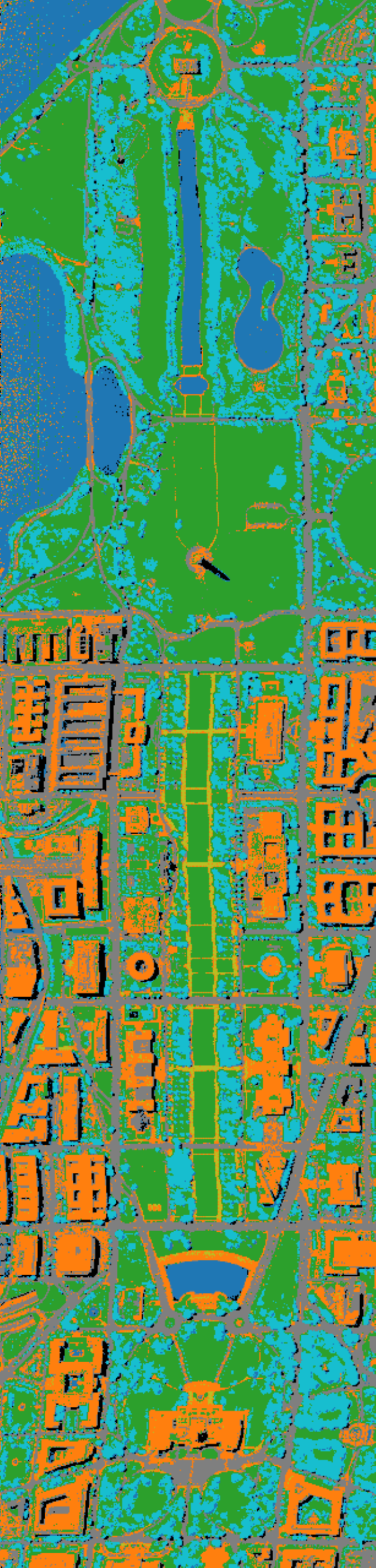}
        \caption{Ground Truth}
    \end{subfigure}\hfill
    \begin{subfigure}[b]{0.135\textwidth}
        \centering
        \includegraphics[width=\linewidth,height=0.25\textheight,keepaspectratio]{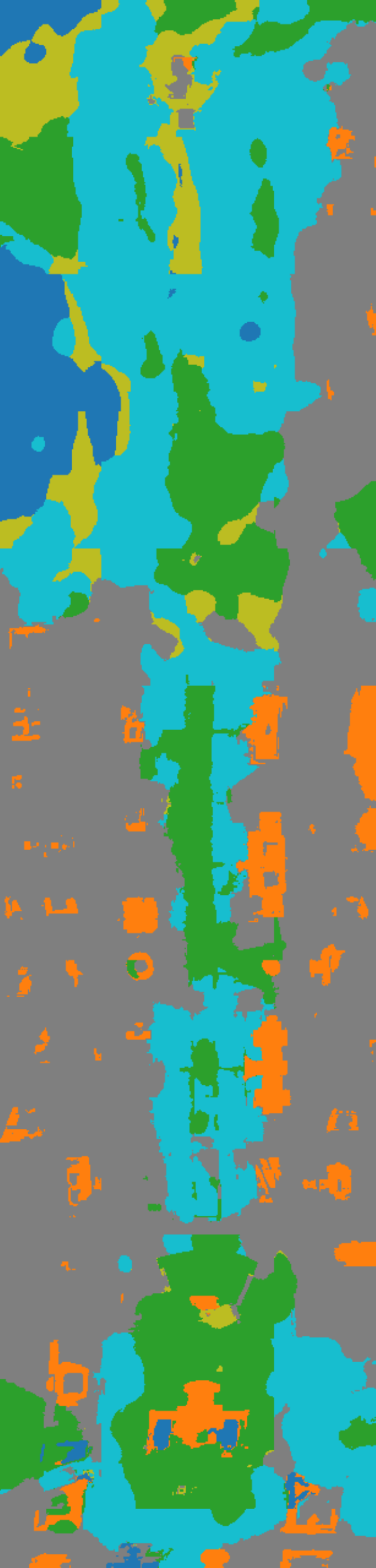}
        \caption{SegEarth-OV}
    \end{subfigure}\hfill
    \begin{subfigure}[b]{0.135\textwidth}
        \centering
        \includegraphics[width=\linewidth,height=0.25\textheight,keepaspectratio]{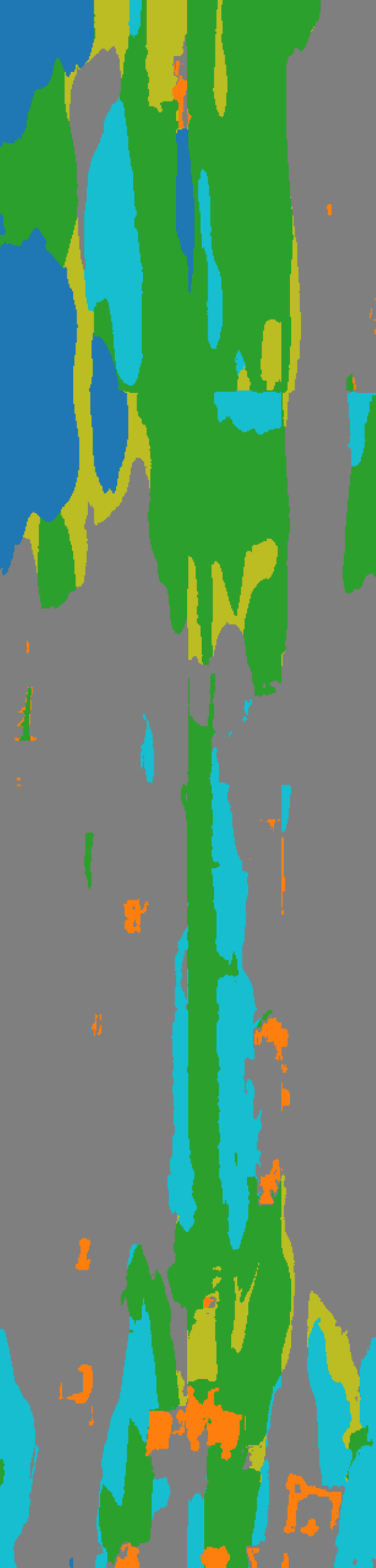}
        \caption{RSCLIP}
    \end{subfigure}\hfill
    \begin{subfigure}[b]{0.135\textwidth}
        \centering
        \includegraphics[width=\linewidth,height=0.25\textheight,keepaspectratio]{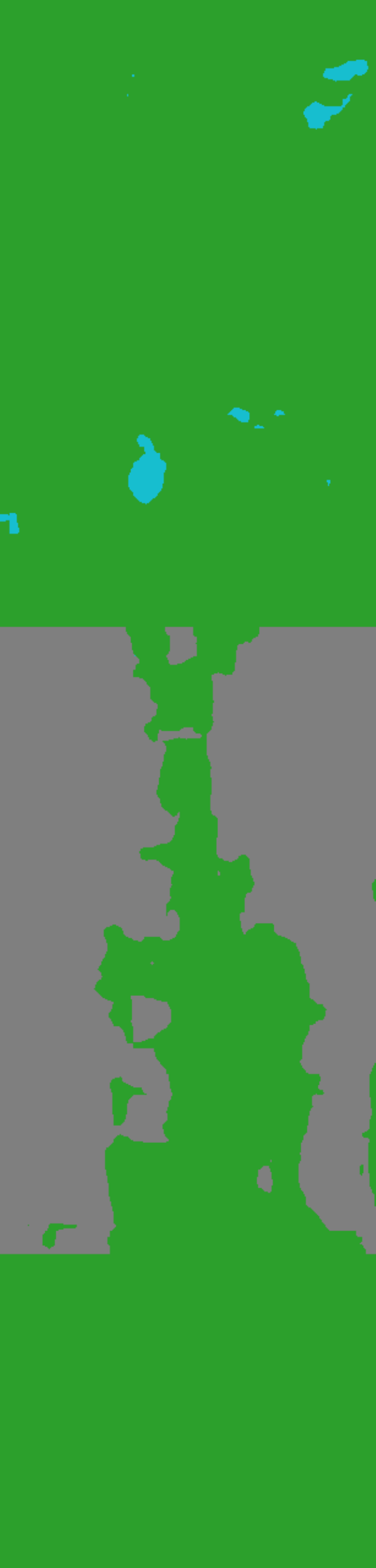}
        \caption{RSKT-Seg}
    \end{subfigure}\hfill
    \begin{subfigure}[b]{0.135\textwidth}
        \centering
        \includegraphics[width=\linewidth,height=0.25\textheight,keepaspectratio]{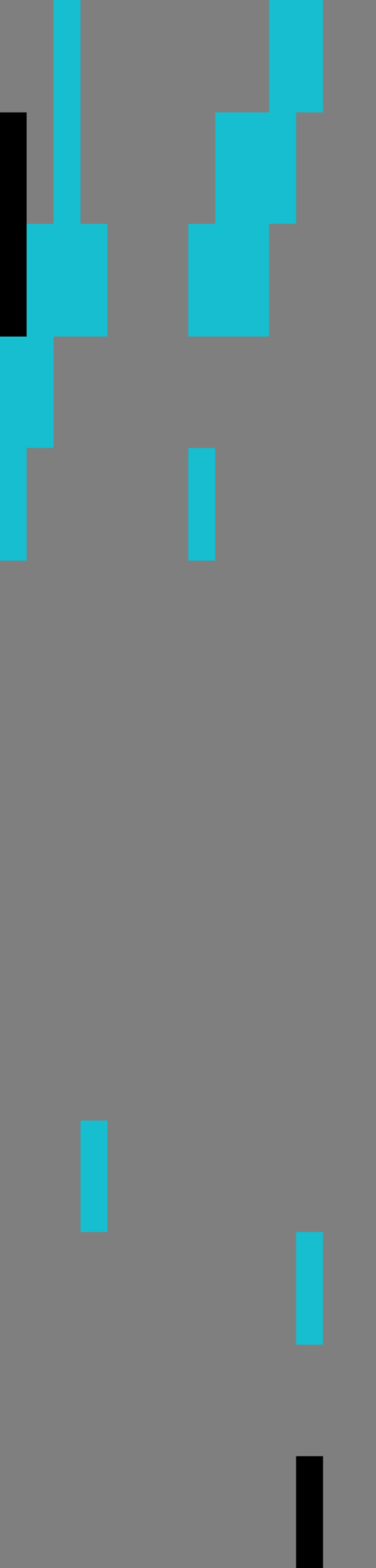}
        \caption{HyperSIGMA}
    \end{subfigure}\hfill
    \begin{subfigure}[b]{0.135\textwidth}
        \centering
        \includegraphics[width=\linewidth,height=0.25\textheight,keepaspectratio]{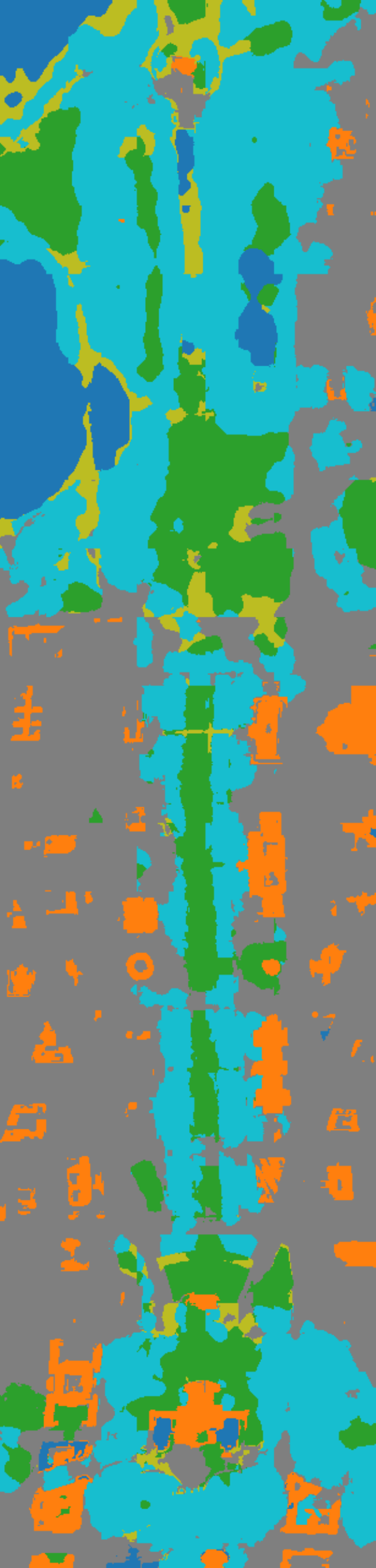}
        \caption{SPECIAL}
    \end{subfigure}\hfill
    \begin{subfigure}[b]{0.135\textwidth}
        \centering
        \includegraphics[width=\linewidth,height=0.25\textheight,keepaspectratio]{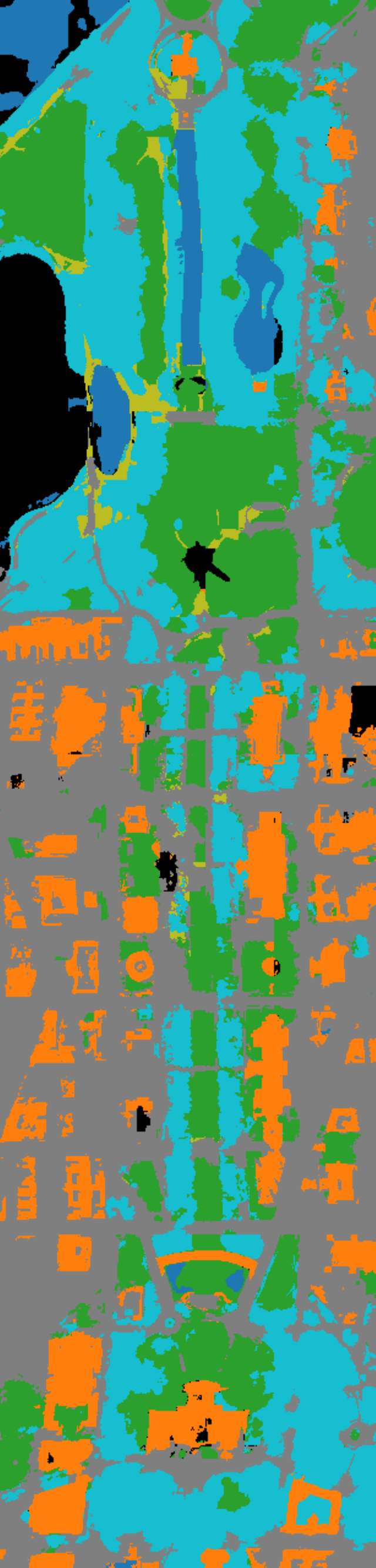}
        \caption{\textbf{Ours}}
    \end{subfigure}

    \vspace{1mm}
   \begin{minipage}{\textwidth}
    \centering
    \small
    \renewcommand{\arraystretch}{1.1}
    \textbf{Legend:}\hspace{0.8em}
    \begin{tabular}{@{\hspace{0.5em}}lllllll@{\hspace{0.5em}}}
    \textcolor[HTML]{000000}{\rule{1.5ex}{1.5ex}} Shadow &
    \textcolor[HTML]{ff7f0e}{\rule{1.5ex}{1.5ex}} Roof &
    \textcolor[HTML]{1f77b4}{\rule{1.5ex}{1.5ex}} Water &
    \textcolor[HTML]{7f7f7f}{\rule{1.5ex}{1.5ex}} Street &
    \textcolor[HTML]{2ca02c}{\rule{1.5ex}{1.5ex}} Grass &
    \textcolor[HTML]{bcbd22}{\rule{1.5ex}{1.5ex}} Path &
    \textcolor[HTML]{17becf}{\rule{1.5ex}{1.5ex}} Trees
\end{tabular}
\end{minipage}
    \vspace{-3mm}
   \caption{Qualitative comparison of segmentation maps on the Washington DC Mall dataset. Previous methods exhibit category confusion in heterogeneous urban regions, whereas SpectralRoute better preserves the spatial organization of major land-cover categories and captures fine-grained structures.}
    \label{fig:supp_washington_vis}
\end{figure*}

To provide a more detailed assessment beyond the overall metrics reported in the main paper, we present the class-wise segmentation accuracy on Pavia University, Washington DC Mall, and Houston 2018 in Tables~\ref{tab:supp_pavia_perclass}, \ref{tab:supp_washington_perclass}, and \ref{tab:supp_houston_perclass}, respectively. These results offer a closer examination of the segmentation performance across land-cover categories with diverse spectral characteristics and semantic properties.

On Pavia University, SpectralRoute achieves an Overall Accuracy (OA) of 70.10\%, an Average Accuracy (AA) of 49.38\%, and a Kappa of 58.21\%. At the category level, it obtains high accuracies on Meadows, Bitumen, Shadows, and Self-Blocking Bricks, reaching 96.99\%, 76.47\%, 59.13\%, and 41.50\%, respectively. It also achieves 93.74\% accuracy on Asphalt, demonstrating its ability to recognize categories with diverse material and spectral characteristics.

On Washington DC Mall, SpectralRoute achieves 54.96\% OA, 41.50\% AA, and 43.60\% Kappa, outperforming all competing methods across the three aggregate evaluation metrics. The class-wise results provide further insight into this improvement. In particular, SpectralRoute achieves a notable advantage on Roof, reaching 53.65\% accuracy compared with 33.52\% for the second-best method. It also maintains competitive performance on Street, Grass, and Trees, indicating that its spectral representations capture discriminative cues across different urban land-cover categories. The resulting improvement in AA further suggests that the gains extend beyond dominant categories, contributing to more balanced and reliable recognition across the evaluated classes.

Houston 2018 encompasses diverse urban land-cover categories, including vegetation, buildings, transportation surfaces, and urban facilities. As shown in Table~\ref{tab:supp_houston_perclass}, SpectralRoute achieves 40.99\% OA, 47.51\% AA, and 32.23\% Kappa, exceeding the best competing results by 5.20\%, 19.91\%, and 8.99\%, respectively. At the class level, it achieves high accuracies on Healthy Grass (97.87\%), Unpaved Parking Lots (97.44\%), Cars (90.34\%), and Stadium Seats (83.32\%). Notably, SpectralRoute successfully recognizes Unpaved Parking Lots, for which all competing methods obtain zero accuracy. It also achieves the best results on Stressed Grass, Bare Earth, and Residential Buildings, highlighting its effectiveness across diverse land-cover types. However, Roads and Sidewalks remain challenging, indicating room for further improvement in fine-grained urban surface discrimination.


Overall, the class-wise results complement the aggregate metrics and demonstrate the consistent performance of SpectralRoute across datasets with different acquisition conditions and class compositions. The category-specific results further illustrate the complementary roles of spectral routing and physics-guided refinement: the former selects relevant spectral views for individual categories, while the latter incorporates spectral consistency into semantic predictions. Together, they support robust open-vocabulary segmentation across diverse hyperspectral scenes.

\subsection{Additional Qualitative Comparisons}

\begin{table*}[ht]
\centering
\caption{Class-wise accuracy comparison of open-vocabulary semantic segmentation methods on the Houston 2018 dataset. \protect\best{Best} and \protect\secondbest{second best} performances are highlighted.}
\label{tab:supp_houston_perclass}
\small
\renewcommand{\arraystretch}{1.1}
\setlength{\tabcolsep}{3pt}
\begin{tabular*}{\textwidth}{@{\hspace{0.8em}\extracolsep{\fill}}lcccccc@{\hspace{0.8em}}}
\toprule
\textbf{Class}
& \textbf{SegEarth-OV~\cite{li2025segearth}}
& \textbf{RSCLIP~\cite{wang2026rsclip}}
& \textbf{RSKT-Seg~\cite{li2026rskt}}
& \textbf{HyperSIGMA~\cite{Wang_2025_HyperSIGMA}}
& \textbf{SPECIAL~\cite{pang2026special}}
& \textbf{SpectralRoute} \\
\midrule
Healthy Grass
& \secondbest{97.55} & 57.31 & 84.69 & 0.00 & 94.73 & \best{97.87} \\

Stressed Grass
& 0.00 & 0.00 & \secondbest{14.29} & 0.00 & 0.00 & \best{22.96} \\

Artificial Turf
& 1.46 & \secondbest{91.12} & \best{100.00} & 0.00 & 0.00 & 45.36 \\

Evergreen Trees
& \best{93.67} & 2.19 & 0.00 & 0.00 & \secondbest{93.13} & 72.70 \\

Deciduous Trees
& 0.02 & 0.00 & \best{26.00} & 0.00 & 0.00 & \secondbest{23.97} \\

Bare Earth
& \secondbest{3.46} & 0.00 & 0.00 & 0.00 & 0.65 & \best{39.58} \\

Water
& \best{68.80} & 0.00 & 0.00 & 0.00 & \best{68.80} & \best{68.80} \\

Residential Buildings
& \secondbest{37.06} & 3.83 & 8.17 & 0.00 & 6.85 & \best{55.19} \\

Non-Residential Buildings
& 22.29 & \secondbest{32.65} & \best{61.20} & 0.00 & 12.45 & 48.48 \\

Roads
& \secondbest{51.79} & 6.46 & 26.97 & 0.00 & \best{64.56} & 17.98 \\

Sidewalks
& \best{12.89} & 0.01 & 2.21 & 0.00 & \secondbest{6.20} & 2.16 \\

Crosswalks
& 0.00 & 0.00 & \best{25.94} & \secondbest{13.96} & 0.00 & 10.23 \\

Major Thoroughfares
& 9.34 & \best{63.27} & 10.18 & 0.00 & 1.90 & \secondbest{26.27} \\

Highways
& 10.39 & \secondbest{71.69} & 18.91 & 0.00 & 10.20 & \best{74.28} \\

Railways
& 0.00 & 0.00 & 0.00 & \best{79.13} & 0.00 & \secondbest{51.50} \\

Paved Parking Lots
& \best{22.60} & \secondbest{13.19} & 1.34 & 0.00 & 7.47 & 15.46 \\

Unpaved Parking Lots
& 0.00 & 0.00 & 0.00 & 0.00 & 0.00 & \best{97.44} \\

Cars
& \secondbest{84.12} & 0.00 & 80.47 & 0.00 & 73.33 & \best{90.34} \\

Trains
& 3.16 & 0.00 & 0.00 & 0.00 & \secondbest{4.35} & \best{6.49} \\

Stadium Seats
& \secondbest{33.34} & 25.48 & 0.00 & 0.00 & 13.36 & \best{83.32} \\
\midrule
OA (\%)
& 26.01 & 24.51 & \secondbest{35.79} & 1.13 & 18.45 & \best{40.99} \\

AA (\%)
& \secondbest{27.60} & 18.36 & 23.02 & 4.65 & 22.90 & \best{47.51} \\

Kappa (\%)
& 17.92 & 9.23 & \secondbest{23.24} & -0.22 & 11.25 & \best{32.23} \\
\bottomrule
\end{tabular*}
\end{table*}

\begin{figure*}[t]
    \centering
    \captionsetup[subfigure]{font=footnotesize,justification=centering}

    \begin{subfigure}[b]{0.135\textwidth}
        \centering
        \includegraphics[width=\linewidth,height=0.25\textheight,keepaspectratio]{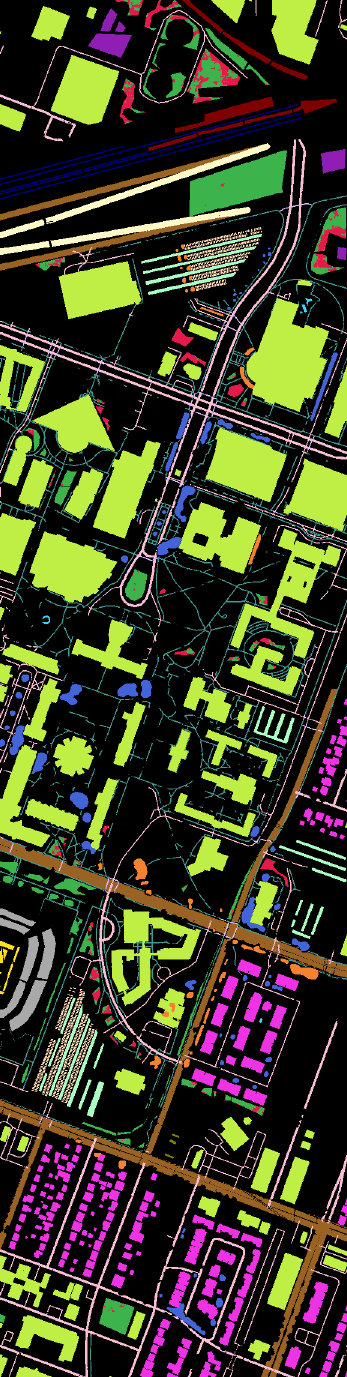}
        \caption{Ground Truth}
    \end{subfigure}\hfill
    \begin{subfigure}[b]{0.135\textwidth}
        \centering
        \includegraphics[width=\linewidth,height=0.25\textheight,keepaspectratio]{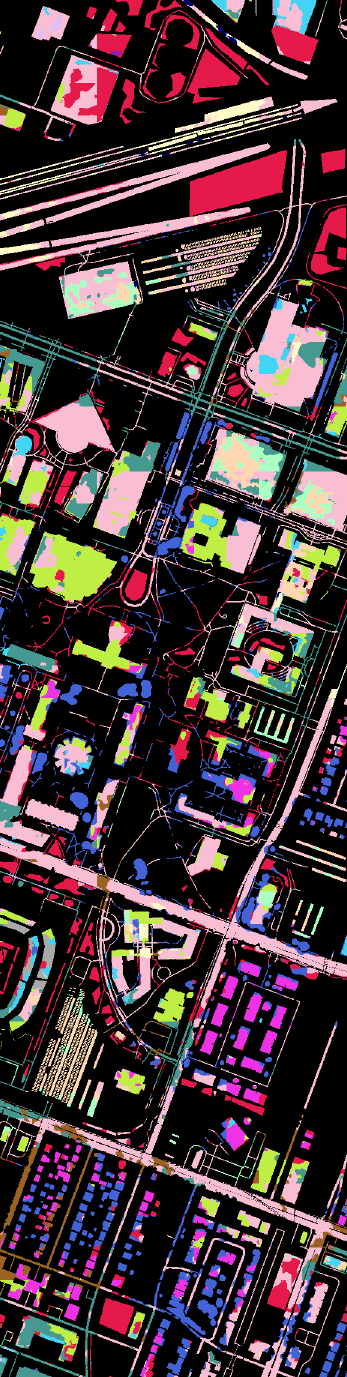}
        \caption{SegEarth-OV}
    \end{subfigure}\hfill
    \begin{subfigure}[b]{0.135\textwidth}
        \centering
        \includegraphics[width=\linewidth,height=0.25\textheight,keepaspectratio]{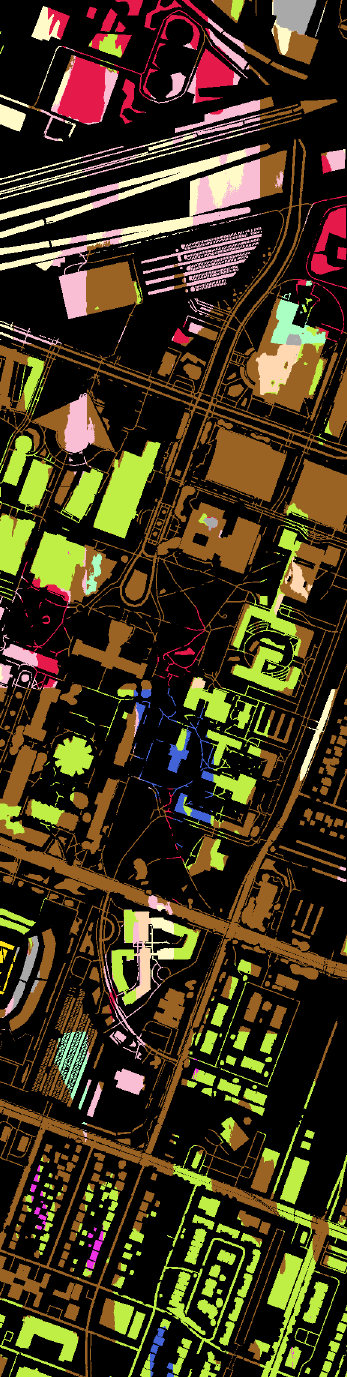}
        \caption{RSCLIP}
    \end{subfigure}\hfill
    \begin{subfigure}[b]{0.135\textwidth}
        \centering
        \includegraphics[width=\linewidth,height=0.25\textheight,keepaspectratio]{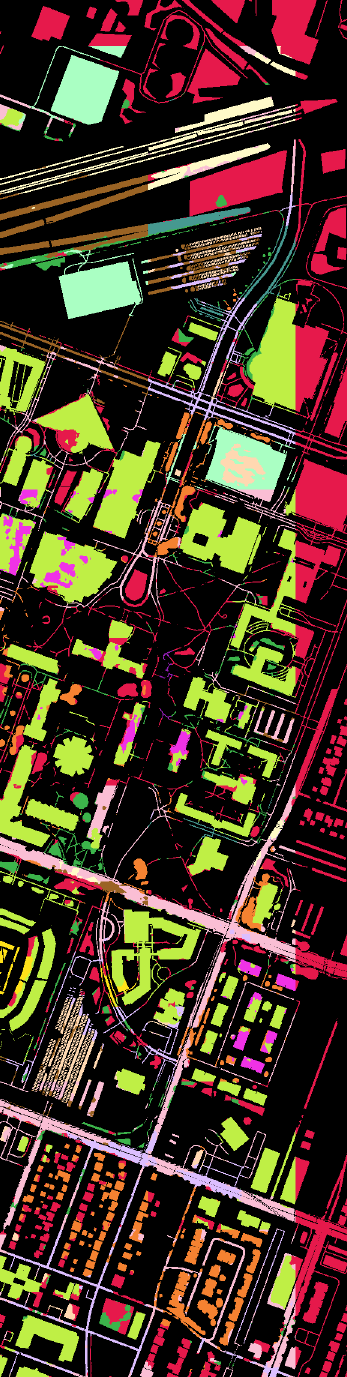}
        \caption{RSKT-Seg}
    \end{subfigure}\hfill
    \begin{subfigure}[b]{0.135\textwidth}
        \centering
        \includegraphics[width=\linewidth,height=0.25\textheight,keepaspectratio]{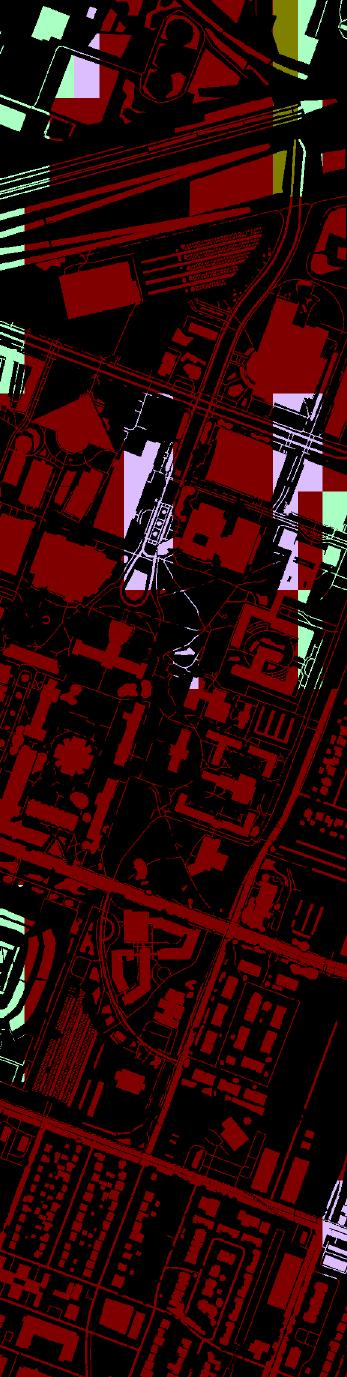}
        \caption{HyperSIGMA}
    \end{subfigure}\hfill
    \begin{subfigure}[b]{0.135\textwidth}
        \centering
        \includegraphics[width=\linewidth,height=0.25\textheight,keepaspectratio]{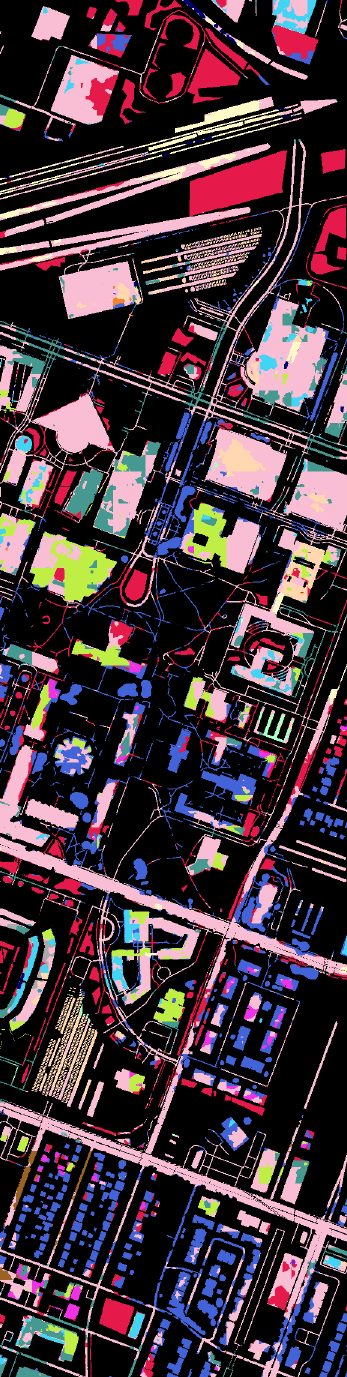}
        \caption{SPECIAL}
    \end{subfigure}\hfill
    \begin{subfigure}[b]{0.135\textwidth}
        \centering
        \includegraphics[width=\linewidth,height=0.25\textheight,keepaspectratio]{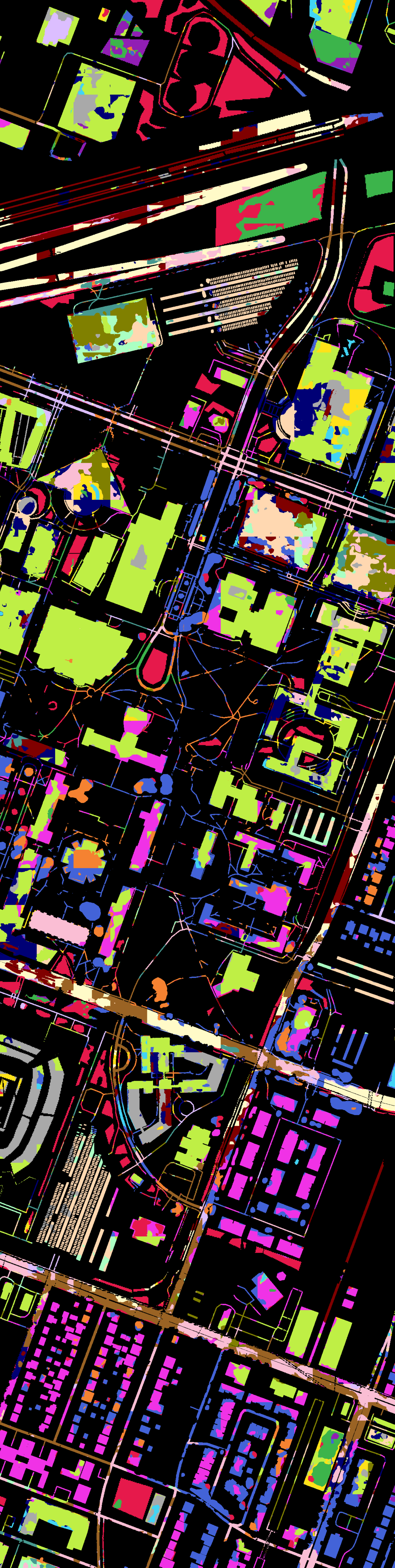}
        \caption{\textbf{Ours}}
    \end{subfigure}

    \vspace{1mm}
    \begin{minipage}{\textwidth}
        \centering
        \scriptsize
        \renewcommand{\arraystretch}{1.1}
        \textbf{Legend:}\hspace{0.5em}
        \begin{tabular}{lllllll}
            \textcolor[HTML]{e6194b}{\rule{1.2ex}{1.2ex}} Healthy Grass &
            \textcolor[HTML]{3cb44b}{\rule{1.2ex}{1.2ex}} Stressed Grass &
            \textcolor[HTML]{ffe119}{\rule{1.2ex}{1.2ex}} Artificial Turf &
            \textcolor[HTML]{4363d8}{\rule{1.2ex}{1.2ex}} Evergreen Trees &
            \textcolor[HTML]{f58231}{\rule{1.2ex}{1.2ex}} Deciduous Trees &
            \textcolor[HTML]{911eb4}{\rule{1.2ex}{1.2ex}} Bare Soil &
            \textcolor[HTML]{42d4f4}{\rule{1.2ex}{1.2ex}} Water \\

            \textcolor[HTML]{f032e6}{\rule{1.2ex}{1.2ex}} Residential Buildings &
            \textcolor[HTML]{bfef45}{\rule{1.2ex}{1.2ex}} Non-Residential Buildings &
            \textcolor[HTML]{fabed4}{\rule{1.2ex}{1.2ex}} Roads &
            \textcolor[HTML]{469990}{\rule{1.2ex}{1.2ex}} Sidewalks &
            \textcolor[HTML]{dcbeff}{\rule{1.2ex}{1.2ex}} Crosswalks &
            \textcolor[HTML]{9a6324}{\rule{1.2ex}{1.2ex}} Major Thoroughfares &
            \textcolor[HTML]{fffac8}{\rule{1.2ex}{1.2ex}} Highways \\

            \textcolor[HTML]{800000}{\rule{1.2ex}{1.2ex}} Railways &
            \textcolor[HTML]{aaffc3}{\rule{1.2ex}{1.2ex}} Paved Parking Lots &
            \textcolor[HTML]{808000}{\rule{1.2ex}{1.2ex}} Unpaved Parking Lots &
            \textcolor[HTML]{ffd8b1}{\rule{1.2ex}{1.2ex}} Cars &
            \textcolor[HTML]{000075}{\rule{1.2ex}{1.2ex}} Trains &
            \textcolor[HTML]{a9a9a9}{\rule{1.2ex}{1.2ex}} Stadium Seats &
            \textcolor[HTML]{000000}{\rule{1.2ex}{1.2ex}} Unlabeled
        \end{tabular}
    \end{minipage}
    \vspace{-3mm}
  \caption{Qualitative comparison of segmentation maps on the Houston 2018 dataset. SpectralRoute demonstrates favorable category discrimination and captures detailed spatial patterns in complex urban scenes.}
    \label{fig:supp_houston_vis}
\end{figure*}

The qualitative results on Washington DC Mall are presented in Fig.~\ref{fig:supp_washington_vis}. 
The scene exhibits distinct spatial patterns, with relatively extensive natural areas in the upper region and densely distributed urban structures in the lower region. 
The segmentation maps illustrate how different methods capture these large-scale land-cover distributions as well as the spatial details within heterogeneous regions. 
SpectralRoute produces predictions that closely follow the overall spatial organization of the ground truth, preserving the distinction between the major natural and built-up areas. 
It also captures a range of fine-grained structures within the urban region, where multiple land-cover categories are spatially interspersed. 
These results highlight the importance of category-specific spectral representation for distinguishing materials that may exhibit similar visual appearances, particularly around transitions between heterogeneous land-cover regions. 
The combination of multi-view spectral routing and physics-guided refinement enables SpectralRoute to exploit complementary spectral evidence while maintaining spatially consistent semantic predictions.

Figure~\ref{fig:supp_houston_vis} presents the qualitative comparison on Houston 2018, which contains complex urban structures with densely distributed buildings, roads, and diverse surface materials. 
Unlike scenes dominated by large homogeneous regions, this dataset involves intricate spatial arrangements and frequent transitions among different urban land-cover categories. 
The segmentation maps reveal the challenges of preserving fine-scale structures while distinguishing spectrally and visually related materials. 
SpectralRoute captures the major spatial organization of the scene, including the road network and the surrounding building blocks, while retaining diverse local land-cover patterns. 
Its predictions show a close correspondence with the ground truth in the arrangement of major urban structures and the distribution of fine-grained regions. 
This visual evidence further illustrates the role of category-adaptive spectral routing in selecting discriminative information from complementary spectral views. 
Moreover, the PGR introduces additional spectral consistency into semantic predictions, helping resolve ambiguities among land-cover categories with related visual characteristics but different spectral responses.

\section{Additional Ablation Studies and Analysis}
\label{sec:supp_analysis}

\subsection{Additional Fusion Strategy Comparison}
\label{sec:supp_fusion}

To further examine the effectiveness of the proposed fusion strategy, we conduct an additional comparison on Pavia University. As shown in Table~\ref{tab:supp_pavia_fusion}, JSR consistently outperforms Uniform Average and Max Confidence across all three evaluation metrics. Notably, the improvement in AA is particularly evident, suggesting that category-adaptive spectral routing better captures the varying spectral requirements of different land-cover categories. The consistent gains in OA and Kappa further demonstrate its effectiveness in exploiting complementary spectral information for pixel-level recognition. These results complement the comparison on Washington DC Mall in the main paper and provide further evidence for the effectiveness of category-specific spectral fusion.

\begin{table}[t]
\centering
\caption{Comparison of different feature fusion strategies on the Pavia University dataset. The results highlight the effectiveness of category-adaptive spectral routing in improving class-wise recognition performance.}
\label{tab:supp_pavia_fusion}
\renewcommand{\arraystretch}{1.1}
\small
\begin{tabular*}{\columnwidth}{@{\hspace{0.8em}\extracolsep{\fill}} lccc @{\hspace{0.8em}}}
\toprule
\textbf{Fusion Strategy} & \textbf{OA (\%)} & \textbf{AA (\%)} & \textbf{Kappa (\%)} \\
\midrule
Uniform Average & 68.18 & 42.32 & 55.43 \\
Max Confidence & 68.39 & 41.78 & 55.14 \\
Ours (JSR) & \textbf{70.10} & \textbf{49.38} & \textbf{58.21} \\
\bottomrule
\end{tabular*}
\end{table}

\subsection{Extended Physics-Guided Refinement Analysis}
\label{sec:supp_pgr}

The main paper examines the effect of Physics-Guided Refinement (PGR) on visually similar categories, including grass, buildings, and parking lots. To further investigate its contribution across a broader range of land-cover types, we provide additional class-wise comparisons on Houston 2018. As shown in Table~\ref{tab:supp_pgr}, PGR brings improvements to various categories beyond those examined in the main paper, providing further evidence of its effectiveness in refining category-level predictions. Notably, the substantial gains on highways and railways demonstrate the effectiveness of spectral refinement in enhancing the recognition of transportation-related categories with distinct material and surface characteristics. Evergreen Trees also exhibits a considerable improvement, indicating that the benefits extend beyond man-made structures to natural land-cover categories. Meanwhile, the gains on crosswalks, trains, and sidewalks further demonstrate the contribution of spectral evidence to distinguishing different urban objects with potentially overlapping semantic appearances. Even for Cars, which already achieves relatively high accuracy without PGR, spectral refinement provides additional improvement, suggesting that it can offer useful discriminative cues even when the initial predictions are already reliable.

These results complement the analysis in the main paper by demonstrating that the benefits of PGR extend across diverse land-cover types rather than being confined to the selected visually similar categories. By incorporating category-specific spectral consistency into semantic prediction refinement, PGR provides complementary physical evidence that helps improve pixel-level recognition across different materials and urban structures.

\begin{table}[t]
\centering
\caption{Additional class-wise accuracy (\%) on Houston 2018, without and with PGR.}
\label{tab:supp_pgr}
\small
\renewcommand{\arraystretch}{1.1}
\begin{tabular*}{\columnwidth}{@{\hspace{0.8em}\extracolsep{\fill}} lcc @{\hspace{0.8em}}}
\toprule
\textbf{Class} & \textbf{w/o PGR} & \textbf{w/ PGR} \\
\midrule
Highways & 32.32 & 74.28\ \textcolor{green!40!black}{\scriptsize(+41.96\,$\uparrow$)} \\
Railways & 20.82 & 51.50\ \textcolor{green!40!black}{\scriptsize(+30.68\,$\uparrow$)} \\
Crosswalks & 0.73 & 10.23\ \textcolor{green!40!black}{\scriptsize(+9.50\,$\uparrow$)} \\
Sidewalks & 1.08 & 2.16\ \textcolor{green!40!black}{\scriptsize(+1.08\,$\uparrow$)} \\
Cars & 87.12 & 90.34\ \textcolor{green!40!black}{\scriptsize(+3.22\,$\uparrow$)} \\
Trains & 1.69 & 6.49\ \textcolor{green!40!black}{\scriptsize(+4.80\,$\uparrow$)} \\
Evergreen Trees & 44.05 & 72.70\ \textcolor{green!40!black}{\scriptsize(+28.65\,$\uparrow$)} \\
\bottomrule
\end{tabular*}
\end{table}

\subsection{Additional Multi-View Spectral Contribution Analysis}

\begin{figure}[t]
    \centering
    \includegraphics[width=\columnwidth]{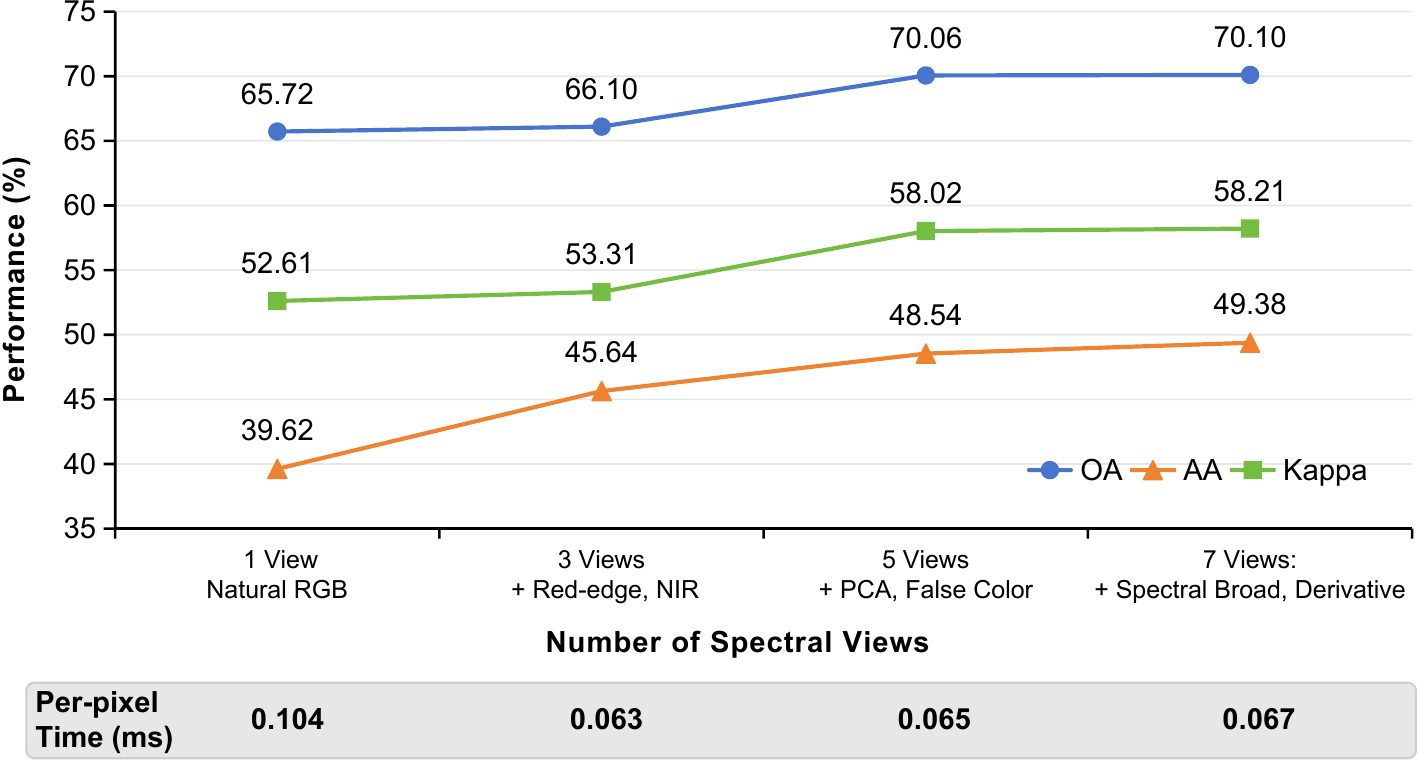}
    \caption{Effect of progressively adding spectral views from spatial appearance to material variations on Pavia University. Per-pixel time is measured after amortizing fixed overheads.}
    \label{fig:pavia_views}
    \vspace{-1em}
\end{figure}

To further examine the contribution of multi-view spectral representations across different scenes, we conduct an additional view ablation on Pavia University. As shown in Fig.~\ref{fig:pavia_views}, the introduction of Red-edge and NIR views leads to a notable improvement in AA, suggesting that the additional spectral cues contribute to more comprehensive category recognition beyond the Natural RGB representation. Incorporating PCA and False Color views further improves all three evaluation metrics, with a particularly clear gain in OA and Kappa. This improvement indicates that the additional spectral representations provide complementary information for distinguishing land-cover categories at the scene level. The final inclusion of Spectral Broad and Derivative views brings further gains, although the improvements become more incremental, suggesting that the seven-view configuration captures complementary spectral information with diminishing additional benefits.

Regarding inference efficiency, as discussed in the main paper, the fixed overheads of the fusion stage are amortized over a larger feature matrix as more spectral views are incorporated. A similar trend is observed on Pavia University: the per-pixel time drops considerably when increasing the number of views from one to three, and remains nearly stable with further view additions. This indicates that the additional views contribute complementary spectral information without substantially increasing the average fusion cost per pixel.

\section{Future Work}
\label{sec:supp_future}

SpectralRoute explores training-free open-vocabulary hyperspectral segmentation by integrating complementary spectral views, knowledge-enhanced text representations, category-adaptive routing, and PGR into a unified framework. While this framework establishes a meaningful connection between spectral observations and semantic concepts, further progress requires more comprehensive modeling of spectral information and a deeper understanding of its relationship with material properties. Several promising directions remain worth exploring.

A natural extension is to move beyond view-based spectral representation toward unified spectral-language modeling. The complementary views adopted in SpectralRoute expose different aspects of hyperspectral observations to frozen VLMs, but they remain transformed representations of the original continuous spectra. Future research could investigate how wavelength-dependent responses and material properties can be directly encoded into a shared spectral-language space. Establishing such associations may provide a richer basis for interpreting material characteristics and recognizing fine-grained categories beyond the predefined vocabulary.

Improving the reliability of spectral evidence is another important direction. Category-adaptive routing identifies spectral views relevant to individual categories, while PGR introduces material-level evidence into semantic predictions. However, the relevance of a spectral view does not necessarily indicate the certainty of its evidence, particularly when distinct materials exhibit overlapping spectral responses. Incorporating uncertainty estimation into routing and PGR could help characterize ambiguous evidence, regulate category-specific corrections, and improve the reliability of fine-grained predictions.

Beyond individual scenes, transferring spectral-semantic knowledge across sensors and geographical regions presents another important challenge. Differences in spectral response functions, band configurations, and acquisition conditions can alter the observed signatures of the same material. Future research could explore physically grounded cross-sensor alignment to distinguish sensor-dependent variations from material-intrinsic characteristics. Establishing more consistent spectral representations across heterogeneous acquisition conditions would provide a basis for extending open-vocabulary hyperspectral interpretation to broader Earth observation scenarios.

\end{document}